\documentclass[runningheads]{llncs}
\usepackage{graphicx}
\usepackage{comment}
\usepackage{amsmath,amssymb} %
\usepackage[utf8]{inputenc}

\usepackage{etoolbox}
\usepackage{xspace}
\usepackage{array,multirow}
\usepackage{boldline}
\usepackage[titlenumbered,ruled ]{algorithm2e}
\usepackage[symbol]{footmisc}
\usepackage{tablefootnote}
\usepackage[flushleft]{threeparttable}
\usepackage{amsmath}
\usepackage{breqn}
\usepackage{tabularx}
\usepackage{wrapfig}
\usepackage[pagebackref=true,breaklinks=true,letterpaper=true,colorlinks,bookmarks=false]{hyperref}
\usepackage[capitalise]{cleveref}

\usepackage{multirow}
\usepackage{verbatim}
\usepackage{color}
\usepackage{float}
\usepackage{enumitem}
\usepackage{booktabs}
\usepackage{tabulary,multirow,overpic,xcolor}
\usepackage[caption=false,font=scriptsize]{subfig}

\usepackage{lipsum}

\usepackage{makecell}

\newcommand{\MyMapTemplatePrefixc}[4]{\expandafter#1\csname#3#4\endcsname{#2{#4}}} 
\forcsvlist{\MyMapTemplatePrefixc {\def} {\mathcal}{c}} {A,B,C,D,E,F,G,H,I,J,K,L,M,N,O,P,Q,R,S,T,U,V,W,X,Y,Z}  
\newcommand{\MyMapTemplatePrefixtb}[5]{\expandafter#1\csname#4#5\endcsname{#2{#3{#5}}}} 
\forcsvlist{\MyMapTemplatePrefixtb {\def} {\tilde}{\mathbf}{t}} {A,B,C,D,E,F,G,H,I,J,K,L,M,N,O,P,Q,R,S,T,U,V,W,X,Y,Z}  
\forcsvlist{\MyMapTemplatePrefixtb {\def} {\tilde}{\mathbf}{t}} {0,1,a,b,c,d,e,f,g,h,i,j,k,l,m,n,o,p,q,r,s,u,v,w,x,y,z}  
\makeatletter
\newcommand\footnoteref[1]{\protected@xdef\@thefnmark{\ref{#1}}\@footnotemark}
\makeatother
\newcommand{\MyMapTemplateNoPrefix}[3]{\expandafter#1\csname#3\endcsname{#2{#3}}}
\forcsvlist{\MyMapTemplatePrefixc {\def} {\mathbf}{mb}} {0,1,a,b,c,d,e,f,g,h,i,j,k,l,m,n,o,p,q,r,s,u,v,w,x,y,z}  
\forcsvlist{\MyMapTemplatePrefixc {\def} {\mathbf}{mb}} {A,B,C,D,E,F,G,H,I,J,K,L,M,N,O,P,Q,R,S,T,U,V,W,X,Y,Z}    

\usepackage{mathtools}

\DeclarePairedDelimiter{\floor}{\lfloor}{\rfloor}

\DeclareMathOperator{\conva}{Conv_A}
\DeclareMathOperator{\convb}{Conv_B}
\DeclareMathOperator{\convc}{Conv_C}
\DeclareMathOperator{\convd}{Conv_D}
\DeclareMathOperator{\shifta}{Shift_1}
\DeclareMathOperator{\shiftc}{Shift_2}
\DeclareMathOperator{\shiftb}{Shift_3}
\DeclareMathOperator{\shiftd}{Shift_4}

\DeclareMathOperator{\conv}{Conv}

\makeatletter
\DeclareRobustCommand\onedot{\futurelet\@let@token\@onedot}
\def\@onedot{\ifx\@let@token.\else.\null\fi\xspace}
\def\eg{\emph{e.g}\onedot} 
\def\ie{\emph{i.e}\onedot} 

\def\etal{\emph{et al}\onedot}
\newcommand\crule[3][black]{\textcolor{#1}{\rule{#2}{#3}}}
\makeatother

\newcolumntype{x}[1]{>{\centering\arraybackslash}p{#1pt}}

\newcommand{\app}{\raise.17ex\hbox{$\scriptstyle\sim$}}

\newlength\savewidth
\newcommand{\tablestyle}[2]{\setlength{\tabcolsep}{#1}\renewcommand{\arraystretch}{#2}\centering\footnotesize}

\newcommand\blfootnote[1]{%
  \begingroup
  \renewcommand\thefootnote{}\footnote{#1}%
  \addtocounter{footnote}{-1}%
  \endgroup
}

\begin{document}
\pagestyle{headings}
\mainmatter
\def\ECCVSubNumber{160}  %

\title{Autoregressive Unsupervised Image Segmentation}
\titlerunning{Autoregressive Unsupervised Image Segmentation}
\author{Yassine Ouali, Céline Hudelot \and  Myriam Tami}
\institute{Université Paris-Saclay, CentraleSupélec, MICS, 91190, Gif-sur-Yvette, France}
\authorrunning{Y. Ouali, C. Hudelot and  M. Tami}
\institute{Université Paris-Saclay, CentraleSupélec, MICS, 91190, Gif-sur-Yvette, France
\email{\{yassine.ouali,celine.hudelot,myriam.tami\}@centralesupelec.fr}}

\maketitle

\begin{abstract}
In this work, we propose a new unsupervised image segmentation approach based on mutual 
information maximization between different constructed views of the inputs.
Taking inspiration from autoregressive generative models
that predict the current pixel from \textit{past} pixels in a raster-scan ordering
created with masked convolutions,
we propose to use different
\textit{orderings} over the inputs using various forms of
masked convolutions to construct different \textit{views} of the data.
For a given input, the model produces a pair of predictions with two valid orderings, and
is then trained to maximize the mutual information between the two outputs.
These outputs can either be low-dimensional features for
representation learning or output clusters corresponding to semantic labels for clustering.
While masked convolutions are used during training, in inference,
no masking is applied and we fall back to
the standard convolution where the model
has access to the full input. 
The proposed method outperforms current state-of-the-art on unsupervised image segmentation.
It is simple and easy to implement, and can be extended to other visual tasks
and integrated seamlessly into
existing unsupervised learning methods requiring different views of the data.
\blfootnote{{\scriptsize Project page: \url{https://yassouali.github.io/autoreg_seg/}}}
\end{abstract}

\section{Introduction}

\begin{figure}[t]
  \centering
\includegraphics[width=\linewidth]{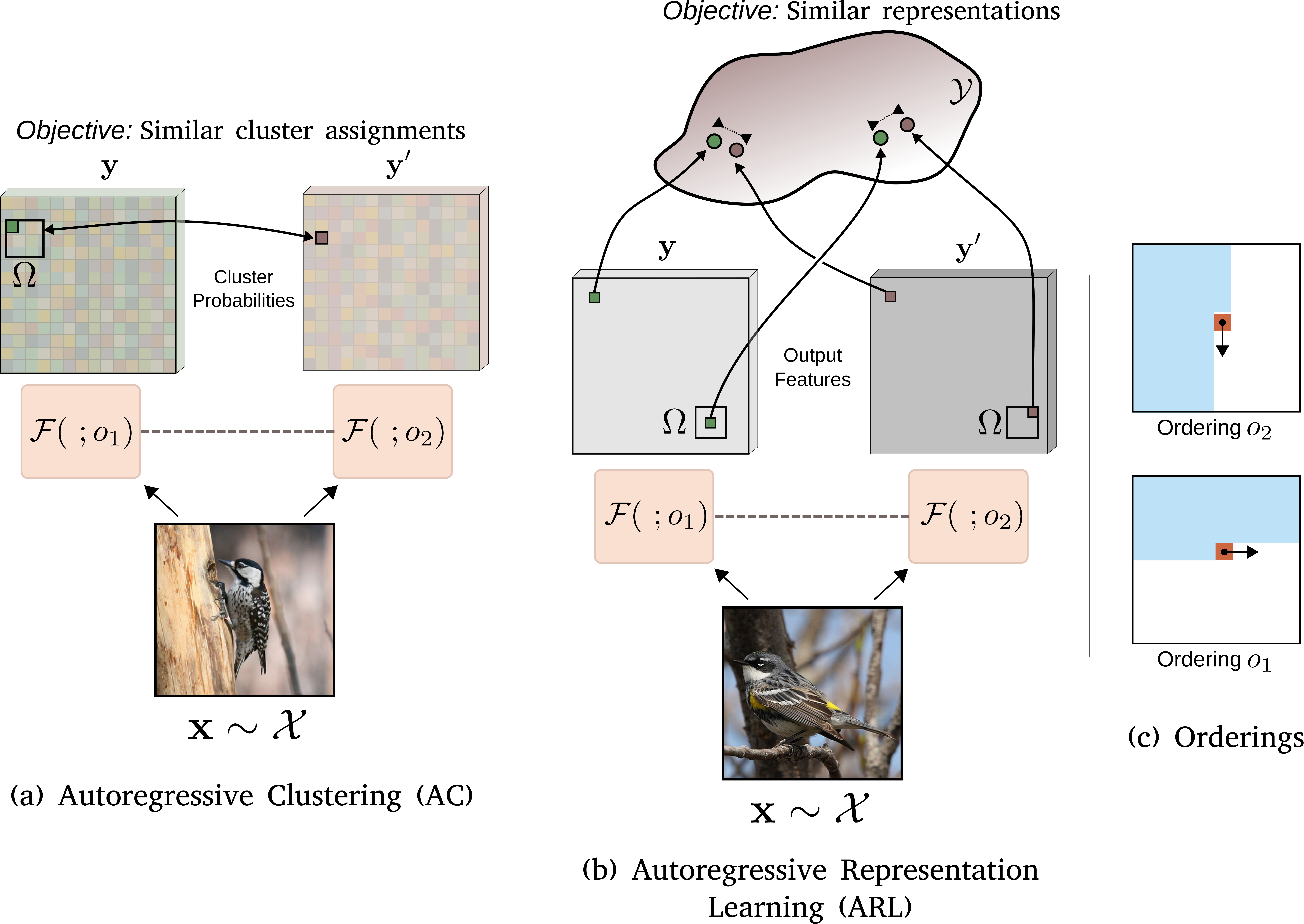}
\vspace{-0.2in}
\caption{\textbf{Overview.} Given an encoder-decoder type
network $\cF$ and two valid orderings $(o_1, o_2)$ as illustrated in (c).
The goal is to maximize the Mutual Information (MI) between
the two outputs over the different
\textit{views}, \ie different \textit{orderings}.
(a) For Autoregressive Clusterings (AC), we output 
the cluster assignments in the form of a probability distribution over pixels, and the goal
is to have similar assignments regardless of the applied ordering.
(b) For Autoregressive Representation Learning (ARL), the objective is to have
similar representations at each corresponding spatial location and its neighbors over a window of small displacements $\Omega$.}
\label{fig:overview}
\vspace{-0.2in}
\end{figure}

Supervised deep learning has enabled great progress and achieved impressive results across a wide number of 
visual tasks, but it requires large annotated datasets for effective training.
Designing such fully-annotated datasets
involves a significant effort in terms of data cleansing and manual labeling.
It is especially true for fine-grained annotations such as pixel-level annotations needed for segmentation tasks,
where the annotation cost per image is considerably high \cite{cocostuff,pascalvoc}.
This hurdle can be overcome with unsupervised learning, where
unknown but useful patterns can be extracted from the easily accessible unlabeled data.
Recent advances in unsupervised learning \cite{infomax,IIC,deepercluster,cpc}, that closed
the performance gap with its supervised counterparts, make it a strong possible alternative. 

Recent works are mainly interested in two objectives, unsupervised representation learning and clustering. 
Representation learning aims to learn
semantic features that are useful for down-stream tasks, be it classification, regression or visualization.
In clustering, the unlabeled data points are directly
grouped into semantic classes. In both cases,
recent works showed the effectiveness of maximizing Mutual Information (MI)
between different \textit{views} of the inputs to 
learn useful and transferable features \cite{infomax,cpc,CMC,federici2020learning} or discover
clusters that accurately match semantic classes \cite{IIC,kANMI}.

Another line of study in unsupervised learning is generative modeling. In particular,
for image modeling, generative autoregressive models \cite{prnn,pcnn,pcnnpp,psnail}, such as PixelCNN,
are powerful generative models with tractable likelihood computation.
In this case, the high-dimensional data, \eg, an image $\mbx$, is factorized as a product
of conditionals over its pixels. The generative
model is then trained to predict the current pixel $x_i$
based on the past values $x_{\leq i-1}$ in a raster scan fashion using
masked convolutions \cite{pcnn} (\cref{fig:convolutions} (a)).

In this work, instead of using a single left to right, top to bottom ordering, we propose
to use several orderings obtained with different forms of masked convolutions and attention mechanism.
The various \textit{orderings} over the input
pixels, or the intermediate representations, are then considered as
different \textit{views} of the input image\footnote{Throughout the paper, a \textit{view} refers to
the application of a given \textit{ordering}. Both are used interchangeably.},
and the model is then trained to
maximize the MI between the outputs over these different views.

Our approach is generic, and can be applied for both clustering and
representation learning (see \cref{fig:overview}).
For a clustering task (\cref{fig:overview} (a)), we apply a pair
of distinct orderings over a given input image,
producing two pixel-level predictions in the form of probability distribution over the semantic classes.
We then maximize the MI between the two outputs at each
corresponding spatial location and its intermediate neighbors.
Maximizing the MI helps avoiding degeneracy (\eg, uniform output distributions)
and trivial solutions (\eg, assigning all of the pixels to the same cluster).
For representation learning (\cref{fig:overview} (b)),
we maximize a lower bound of MI between the two output feature maps over the different \textit{views}.

We evaluate the proposed method using standard image segmentation
datasets: Potsdam \cite{potsdam} and COCO-stuff \cite{cocostuff}, and show competitive results. We present an extensive
ablation study to highlight the contribution of each component within the proposed framework, 
and emphasizing the flexibility of the method.

To summarize, we propose following contributions:
\textbf{(i)} a novel unsupervised method for image segmentation based on autoregressive models and MI maximization;
\textbf{(ii)} various forms of masked convolutions to generate different orderings;
\textbf{(iii)} an attention augmented version of masked convolutions for a larger receptive
field, and a larger set of possible orderings;
\textbf{(iv)} an improved performance above previous state-of-the-art on unsupervised image segmentation.

\section{Related Works}

\noindent\textbf{Autoregressive models.}
Many autoregressive models
\cite{NADE,MADE,pcnn,pcnnpp,imtransformer,psnail,sparsetransformers}
for natural image modeling have been proposed.
They model the joint probability distribution
of high-dimensional images as a product of conditionals over the pixels.
PixelCNN \cite{pcnn,prnn} specifies the conditional distribution
of a sub-pixel (\ie, a color channel of a pixel) as a full 256-way softmax,
while PixelCNN++ \cite{pcnnpp} uses 
a mixture of logistics. In both cases, masked convolutions are used to
process the initial image $\mbx$ in an autoregressive manner.
In Image \cite{imtransformer} and Sparse \cite{sparsetransformers} transformers,
self-attention \cite{attalluneed} is used over the input pixels, while PixelSNAIL \cite{psnail}
combines both attention and masked convolutions.%

\noindent\textbf{Clustering and unsupervised representation learning.}
Recent works in clustering aim at combining traditional clustering algorithms
\cite{clustering} with deep learning,
such as using K-means style objectives when training deep nets
training \cite{deepcluster,associativeclustering,deepkmeans}.
However, such objective can lead to trivial and degenerate solutions \cite{deepcluster}.
IIC \cite{IIC} proposed to use a MI based objective which is intrinsically
more robust to such trivial solutions. 
Unsupervised learning of representations
\cite{infomax,amdim,cpc,imrot} rather aims to train a model,
mapping the unlabeled inputs into some
lower-dimensional space, while preserving semantic information and discarding instance-specific details.
The pre-trained model can then be fine-tuned on a down-stream task with fewer labels.

\noindent\textbf{Unsupervised learning and MI maximization.}
Maximizing MI for unsupervised learning is not a new idea
\cite{clustering,becker1992self}, and recent works demonstrated its
effectiveness for unsupervised learning.
For representation learning, the training objective is to maximize
a lower bound of MI over continuous random variables 
between distinct views of the inputs.
These views can be the input image and its representation
\cite{IMSAT}, the global and local
features \cite{infomax}, the features at different scales \cite{amdim},
a sequence of extracted patches from an image
in some fixed order \cite{cpc} or different modalities of the image \cite{CMC}.
For a clustering objective, with discrete random variables as outputs,
the exact MI can be maximized over the different views,
\eg, IIC \cite{IIC} maximizes the MI between the image and its augmented version.

\noindent\textbf{Unsupervised Image Segmentation.}
Methods that learn the segmentation masks entirely from data with no supervision can be categorized as follows: (1) GAN based methods \cite{redo,bielski2019emergence} that extract and redraw the main object in the image for object segmentation. Such methods are limited to only instances with two classes, a foreground and a background. The proposed method is more generalizable and is independent of the number of ground-truth classes; (2) Iterative methods \cite{segsort} consisting of a two-step process. The features produced by a CNN are first grouped into clusters using spherical K-means. The CNN is then trained for better feature extraction to discriminate between the clusters. We propose an end-to-end method  simplifying both training and inference; (3) MI maximization based methods \cite{IIC} where the MI between two views of the same instance at the corresponding spatial locations is maximized. We propose an efficient and effective way to create different views of the input using masked convolutions. Another line of work consists of leveraging the learned representations of a deep network for unsupervised segmentation, \eg, CRFs \cite{wnet} and deep priors \cite{kanezaki2018unsupervised}.

\section{Method}

Our goal is to learn a representation that maximizes the MI, denoted as $I$,
between different views of the input. These views are generated using various orderings,
capturing different aspects  of the inputs.
Formally, let $\mbx \sim \cX$ be an unlabeled data point, and
$\cF : \cX \rightarrow \cY$ be a deep representation to be learned as a mapping
between the inputs and the outputs.
For clustering, $\cY$ is the set of possible
clusters corresponding to semantic 
classes, and for representation learning, $\cY$ corresponds to a lower-dimensional
space of the output features. Let $(o_i, o_j) \in \cO$ be
two orderings $o_i$ and $o_j$ obtained from the set of possible and valid 
orderings $\cO$ (\cref{fig:raster_orderings}).
For two outputs $\mby \sim \cF(\mbx; o_i)$ and $\mby^{\prime} \sim \cF(\mbx; o_j)$,
the objective is to maximize the predictability of $\mby$ from
$\mby^{\prime}$ and vice-versa, where $\cF(\mbx; o_i)$ corresponds to applying the learning function $\cF$
with a given ordering $o_i$ to process the image $\mbx$.
This objective is equivalent to maximizing the MI
between the two encoded variables:
\begin{equation}
\label{eq:mainobj}
\max _{\cF} I(\mby; \mby^{\prime})
\end{equation}

We start by presenting different forms of masked convolutions
to generate various raster-scan orderings,
and propose an attention augmented variant
for a larger receptive field and use it to extend
the set of possible orderings (\cref{orderings}).
We then formulate the training objective for maximizing \cref{eq:mainobj}
for both clustering and unsupervised representation learning (\cref{obj}).
We finally conclude with a flexible design architecture
for the function $\cF$ (\cref{mdl}).

\subsection{Orderings} \label{orderings}

\subsubsection{Masked Convolutions}

\begin{figure}[t]
\centering
\includegraphics[width=\linewidth]{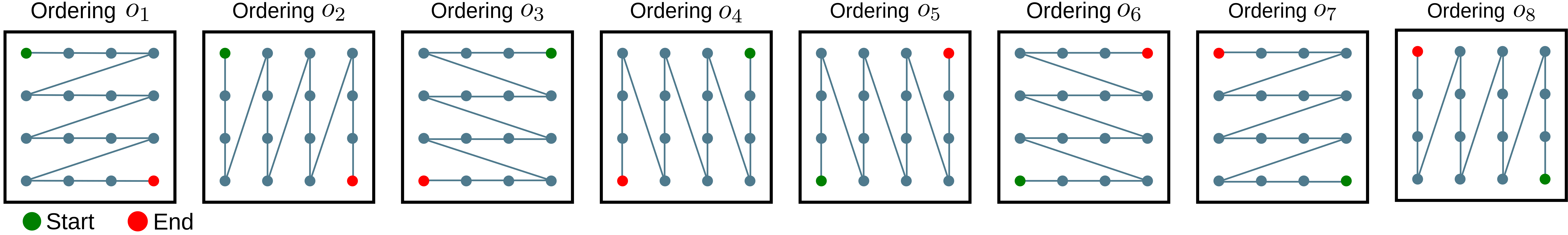}
\vspace{-0.3in}
\caption{\textbf{Raster-scan type orderings.}}
\label{fig:raster_orderings}
\vspace{-0.2in}
\end{figure}

\begin{figure}[t]
\centering
\includegraphics[width=\linewidth]{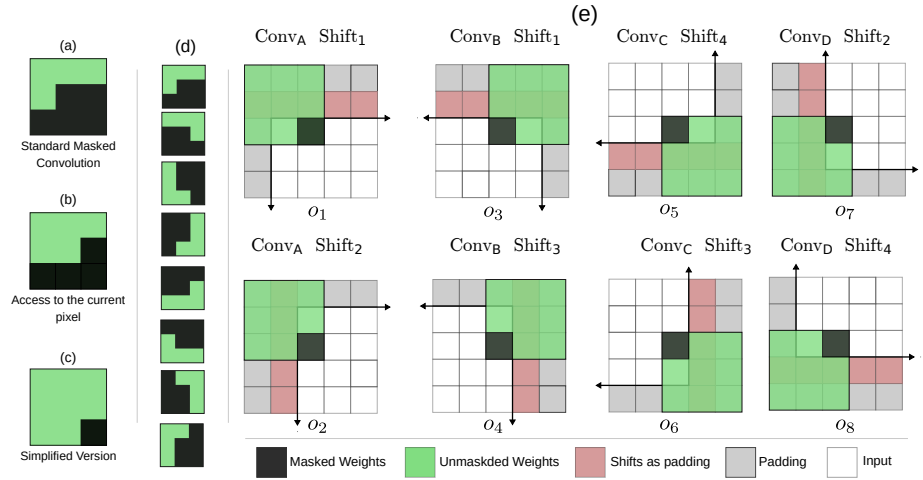}
\vspace{-0.2in}
\caption{\textbf{Masked Convolutions.} (a) Standard masked convolution used in autoregressive
generative modeling, yielding an ordering $o_1$. (b) A relaxed version of standard masked convolution
where we have access to the current pixel at each step.
(c) A simplified version of masked convolution with a reduced number  of masked weights.
(d) The 8 versions of the standard masked convolution to construct all of the possible
raster-scan type orderings.
(e) The proposed types of masked convolutions with
the corresponding shifts to obtain all of the 8 desired raster-scan types orderings.
$F = 3$ in this case.}
\label{fig:convolutions}
\vspace{-0.2in}
\end{figure}

In neural autoregressive modeling \cite{pcnn,pcnnpp,psnail}, for an input image $\mbx \in \mathbb{R}^{H \times W \times 3}$
with $3$ color channels, a raster-scan
ordering is first imposed on the image (see \cref{fig:raster_orderings}, ordering $o_1$).
Such an ordering, where the pixel $x_i$
only depends on the pixels that come before it,
is maintained using masked convolutions\footnote{
Note that for a convolution weight tensor of shape $[F, F, C_{in}, C_{out}]$, the masking in applied over all
values of both channels, $C_{in}$ and $C_{out}$.}
\cite{pcnn,prnn} (\cref{fig:convolutions} (a)).

Our proposition is to use all 8 possible raster-scan type orderings as the set
of valid orderings $\cO$ as illustrated in \cref{fig:raster_orderings}.
A simple way to obtain them is to use a single ordering $o_1$ with the
standard masked convolution (\cref{fig:convolutions} (a)),
along with geometric transformations $g$ (\ie, image rotations by
multiples of 90 degrees and horizontal flips), resulting in 8 versions of the input image.
We can then maximize the MI between the
two outputs, \ie, $I(\mby; g^{-1}(\mby^{\prime}))$ with $\mby^{\prime} \sim \cF(g(\mbx); o_j)$.
In this case, since the masked weights are never trained, 
we cannot fall-back to the normal convolution
where the function $\cF$ has access to the full input during inference,
greatly limiting the performance of such approach.

This point motivates our approach.
Our objective is to learn all the weights
of the masked convolution during training, and use
an unmasked version during inference.
This can be achieved by using a normal convolution,
and for a given ordering $o_i$,
we mask the corresponding weights during the forward
pass to construct the desired view of the inputs.
Then in the backward pass, we only
update the unmasked weights and the masked weights remain unchanged.
In this case, all of the weights will be learned and we will converge to
a normal convolution given enough training iterations.
During inference, no masking is applied, giving
the function $\cF$ full access to the inputs.

A straight forward way to implement this is to use 8 versions of the standard masked
convolution to create the set $\cO$ (\cref{fig:convolutions} (d)).
However, for each forward pass,
the majority of the weights are masked, resulting in a reduced receptive field and a fewer number of weights will be learned at each iteration,
leading to some disparity between them. 

Given that we are interested in a discriminative task, rather than
generative image modeling where the access to the current pixel is not allowed.
We start by relaxing the conditional dependency, and allow the model to have
access to the current pixel, reducing the number of masked locations
by one (\cref{fig:convolutions} (b)).
To further reduce the number of masked weights,
for an $F \times F$ convolution, instead of 
masking the lower rows,
we can simply shift the input by the same amount and only
mask the weights of the last row.
We thus reduce the number of masked weight from
$\floor{F^2 / 2}$ (\cref{fig:convolutions} (b))
to $\floor{F / 2}$ (\cref{fig:convolutions} (c)).
With four possible masked convolutions: $\{\conva, \convb, \convc, \convd\}$
and four possible shifts:\footnote{\eg, for $\shifta$ and a $3 \times 3$ convolution,
an image of spatial dimensions $H \times W$ is first padded on the top resulting in $(H+1) \times W$,
the last row is then cropped, going back to $H \times W$.} $\{\shifta, \shiftb, \shiftc, \shiftd\}$,
we can create all of 8 raster-scan orderings as illustrated in \cref{fig:convolutions} (e).
The proposed masked convolutions do not introduce any additional computational overhead, neither in training,
nor inference, making them easy to implement and integrate into existing architectures with minor changes.

\vspace{-0.1in}
\subsubsection{Attention Augmented Masked Convolutions} \label{acc}
As pointed out by \cite{pcnn}, the proposed masked convolutions
are limited in terms of expressiveness since they create
blind spots in the receptive field (\cref{fig:blind_spots}).
In our case, by applying different orderings, we will have access to all of the input $\mbx$ over
the course of training, and
this \textit{bug} can be seen as a \textit{feature} where the blind spots can be considered as an additional restriction.
This restricted receptive filed, however, can be overcome using the self-attention mechanism \cite{attalluneed}.
Similar to previous works \cite{nonlocal,atgan,attaugmentedconv},
we propose to add attention blocks to model long range dependencies that are hard to access
through standalone convolutions.
Given an input tensor of shape $(H, W, C_{in})$, after reshaping it into a matrix $X \in \mathbb{R}^{HW \times C_{in}}$,
we can apply a masked version of attention \cite{attalluneed} in a straight forward manner.
The output of the attention operation is:
\begin{equation}
\label{eq:att}
A =\operatorname{Softmax}((Q K^{\top}) \odot \mbM_{o_i}) V
\end{equation}
with $Q = XW_q$, $K = XW_k$ and $V = XW_v$, where $W_q, W_k \in \mathbb{R}^{C_{in} \times d}$
and $W_{v} \in \mathbb{R}^{C_{in} \times d}$
are learned linear transformations that map the input $X$ to queries $Q$, keys $K$ and values $V$, and
$\mbM_{o_i} \in  \mathbb{R}^{HW \times HW}$ corresponds to a masking operation
to maintain the correct ordering $o_i$.

\begin{figure}[t]
\centering
\includegraphics[width=\linewidth]{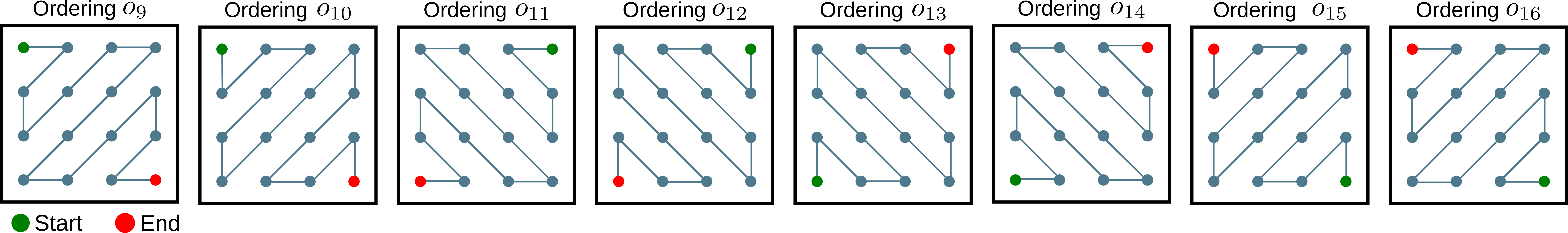}
\vspace{-0.3in}
\caption{\textbf{Zigzag type orderings.}}
\label{fig:zigzag_orderings}
\vspace{-0.2in}
\end{figure}

The output is then projected into the output space using 
a learned linear transformation $W^O \in \mathbb{R}^{d \times C_{in}}$ obtaining $X_{\text{att}} = A W^{O}$.
The output of the attention operation $X_{\text{att}}$ is concatenated channel wise
with the input $X$, and then merged using a $1 \times 1$ convolution
resulting in the output of the attention block.

\vspace{-0.1in}
\subsubsection{Zigzag Orderings.} \label{zo}
\begin{figure}
    \centering
    \begin{minipage}{0.45\textwidth}
        \centering
        \includegraphics[width=0.9\linewidth]{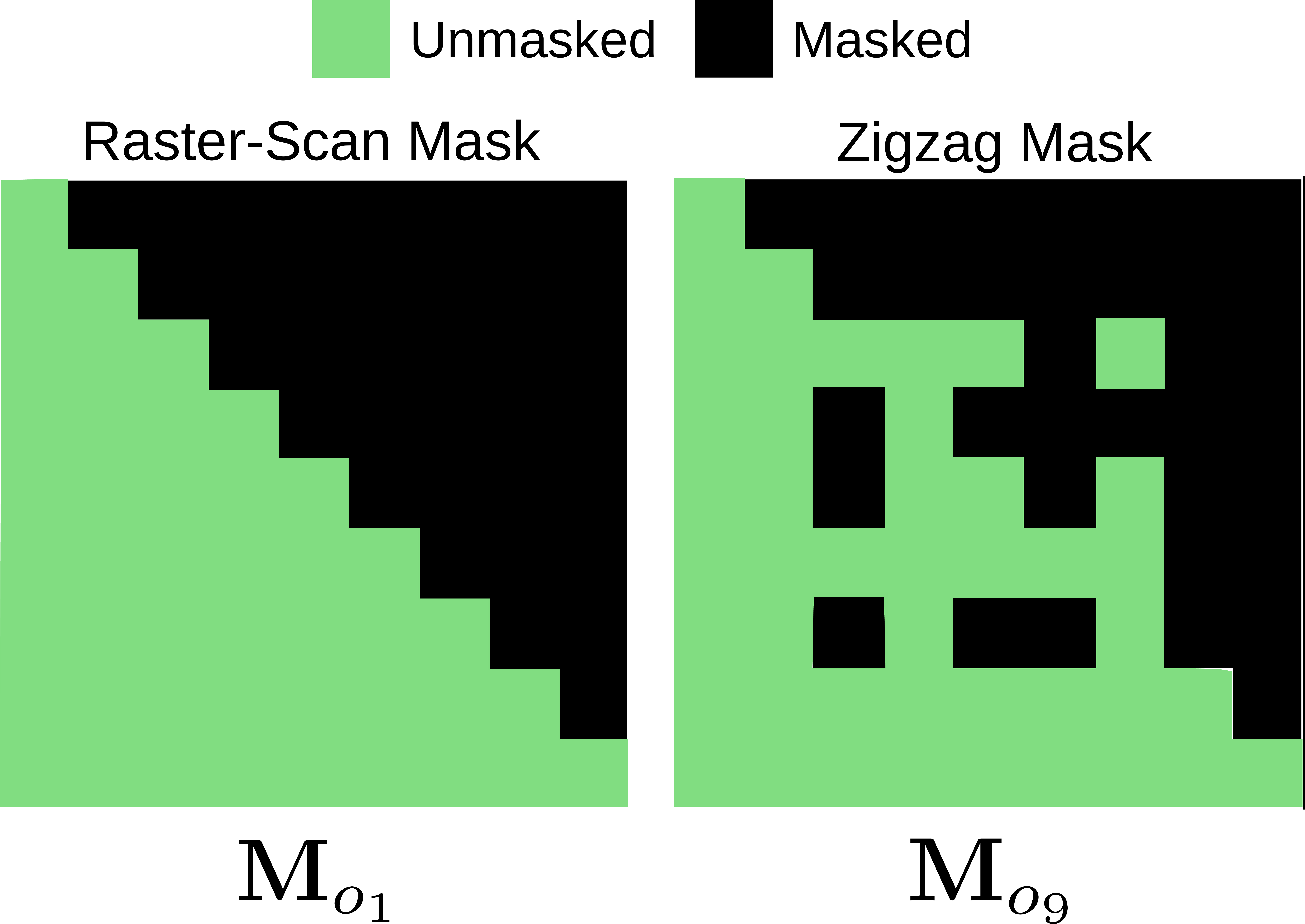}
        \caption{\textbf{Attention Masks.} Examples of the different
          attention masks $\mbM_{o_i}$ of shape $HW \times HW$ applied for a given ordering $o_i$. With $HW = 9$.}
        \label{fig:att_masks}
    \end{minipage}%
    \qquad
    \begin{minipage}{0.45\textwidth}
        \centering
        \includegraphics[width=0.9\linewidth]{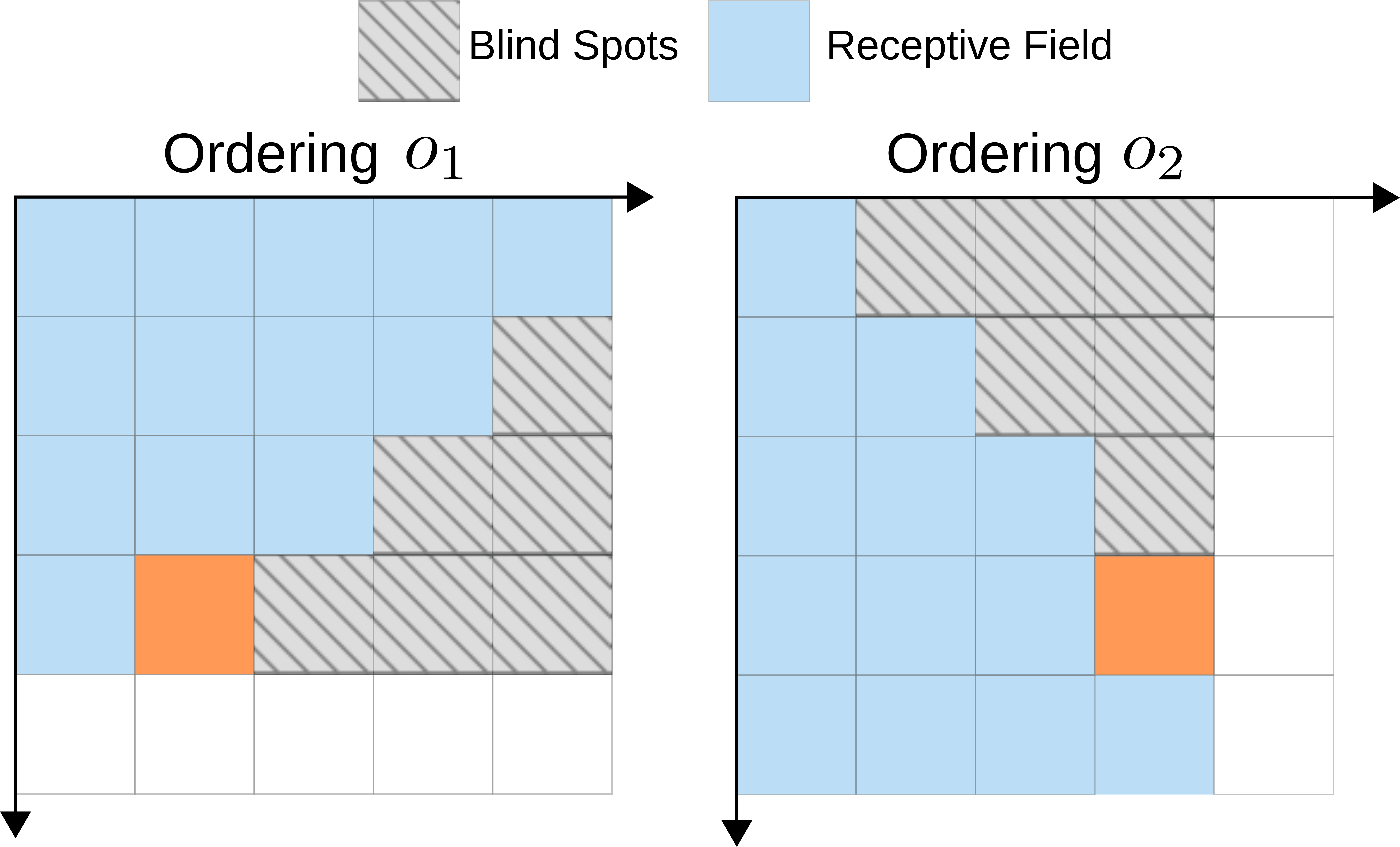}
        \caption{\textbf{Blind Spots.} Blind spots in the receptive field of
        pixel \crule[orange]{0.2cm}{0.2cm} as a result of
        using a masked convolution for a given ordering $o_i$.}
        \label{fig:blind_spots}
    \end{minipage}
\vspace{-0.1in}
\end{figure}

Using attention gives us another benefit, we can extend the set of possible orderings to include
zigzag type orderings introduced in \cite{psnail} (\cref{fig:zigzag_orderings}).
With zigzag orderings, the outputs at each spatial location will be mostly influenced
by the values of the corresponding
neighboring input pixels, which can give rise to more semantically meaningful representations compared
to that of raster-scan orderings. This is done by simply 
using a mask $\mbM_{o_i}$ corresponding to the desired zigzag ordering $o_i$.
Resulting in a set $\cO$ of 16 possible and valid orderings $o_i$ with $i \in \{1, \ldots, 16\}$ in total.
See \cref{fig:att_masks} for an example.

\subsection{Training Objective} \label{obj}
In information theory, the MI $I(X;Y)$ between two random variables $X$ and $Y$
measures the \textit{amount of information} learned from the knowledge of $Y$ 
about $X$ and vice-versa. The MI can be expressed as the difference of two entropy terms:
\begin{equation}
\label{eq:midef}
I(X;Y)=H(X)-H(X|Y)=H(Y)-H(Y|X)
\end{equation}
Intuitively, $I(X; Y)$ can be seen as the reduction of uncertainty in one of the 
variables, when the other one is observed. If $X$ and $Y$ are independent,
knowing one variable exposes nothing about the other, in this case, $I(X; Y) = 0$.
Inversely, if the state of one variable is deterministic when the state of the other is revealed,
the MI is maximized. Such an interpretation explains the goal behind maximizing
\cref{eq:mainobj}.
The neural network $\cF$ must be able to preserve information
and extract semantically similar representations regardless of the applied ordering $o_i$,
and learn representations that encode the underlying shared information between the different views.
The objective can also be interpreted as having a regularization effect,
forcing the function $\cF$ to focus 
on the different views and subparts of the input $\mbx$ to produce similar outputs,
reducing the reliance on specific objects or parts of the image.

Let $p(\mby, \mby^{\prime})$ be the joint distribution produced by
sampling examples $\mbx \sim \cX$ and
then sampling two outputs $\mby \sim \cF(\mbx; o_i)$ and $\mby^{\prime} \sim \cF(\mbx; o_j)$
with two possible orderings $o_i$ and $o_j$.
In this case, the MI in \cref{eq:mainobj} can be defined as the Kullback–Leibler (KL) divergence between
the joint and the product of the marginals:
\begin{equation}
\label{eq:mikl}
I(\mby, \mby^{\prime}) = D_{\mathrm{KL}}(p(\mby, \mby^{\prime})
\| p(\mby) p(\mby^{\prime}))
\end{equation}

To maximize \cref{eq:mikl}, we can either maximize the exact MI for a clustering task over discrete predictions,
or a lower bound for an unsupervised learning of representations over the continuous outputs.
We will now formulate the loss functions $\cL_{\text{AC}}$ and $\cL_{\text{ARL}}$
of both objectives for a segmentation task.

\subsubsection{Autoregressive clustering (AC).} 
In a clustering task, the goal is to train a neural network $\cF$ to predict a
cluster assignment corresponding to a given semantic class $k \in \{1, \ldots, K\}$ with $K$ possible clusters at
each spatial location. 
In this case, the encoder-decoder type 
network $\cF$ is terminated with $K$-way softmax, outputting
$\mby \in [0, 1]^{H \times W \times K}$ of the same spatial dimensions as the input.
Concretely, for a given input image $\mbx$ and two valid orderings $(o_i, o_j) \in \cO$,
we forward pass the input through the network
producing two output probability distributions $\cF(\mbx; o_i) = p(\mby|\mbx, o_i)$ and
$\cF(\mbx; o_j) = p(\mby^{\prime}|\mbx, o_j)$ over the $K$ clusters and at
each spatial location.
After reshaping the outputs into two matrices of shape $HW \times K$, with each element
corresponding to the probability of assigning pixel $x_l$ with $l \in \{1, \ldots, HW\}$ to cluster $k$,
we can compute the joint distribution $p(\mby, \mby^{\prime})$ of shape $K \times K$ as follows:
\begin{equation}
\label{eq:joint}
p(\mby, \mby^{\prime}) = \cF(\mbx; o_i)^{\top} \cF(\mbx; o_j)
\end{equation}
The marginals $p(\mby)$ and $p(\mby^{\prime})$ can then be obtained by
summing over the rows and columns of $p(\mby, \mby^{\prime})$.
Similar to IIC \cite{IIC}, we symmetrize $p(\mby, \mby^{\prime})$ using
$[p(\mby, \mby^{\prime}) + p(\mby, \mby^{\prime})^{\top}] / 2$ to
maximize the MI in both directions.
The clustering loss $\cL_{AC}$ in this case can be written as follows:
\begin{equation}
\label{eq:Lac}
\cL_{\text{AC}} = \mathbb{E}_{\mbx \sim \cX}\left[\mathbb{E}_{p(\mby, \mby^{\prime})}
\log \frac{p(\mby, \mby^{\prime})}{p(\mby) p(\mby^{\prime})}\right]
\end{equation}

In practice, instead of only maximizing the MI between two corresponding spatial locations,
we maximize it between each spatial location and its intermediate neighbors over small
displacements $\mbu \in \Omega$ (see \cref{fig:overview}). This can be efficiently implemented
using a convolution operation as demonstrated in \cite{IIC}.

\subsubsection{Autoregressive representation learning (ARL).} \label{arl}
Although the clustering objective in \cref{eq:Lac} can also
be used as a pre-training objective for $\cF$, Tschannen \etal
\cite{onMImaximization} recently showed  that maximizing 
the MI does not often results in transferable and semantically meaningful features, especially when
the down-stream task is a priori unknown.
To this end, we follow recent representation learning works based on MI maximization
\cite{cpc,infomax,amdim,CMC}, where a lower bound estimate of MI
(\eg, InfoNCE \cite{cpc}, NWJ \cite{NWJ}) is maximized between different views
of the inputs. These estimates are based on the simple intuitive idea, that
if a critic $f$ is able to differentiate between samples drawn from
the joint distribution $p(\mby, \mby^{\prime})$ and samples drawn from the marginals $p(\mby) p(\mby^{\prime})$,
then the true MI is maximized.
We refer the reader to \cite{onMImaximization} for a detailed discussion.

In our case, with image segmentation as the target down-stream task, 
we maximize the InfoNCE estimator \cite{cpc} over the continuous outputs.
Specifically, with two outputs
$(\mby, \mby^{\prime}) \in \mathbb{R}^{H \times W \times C}$ as $C$-dimensional feature maps.
The training objective is to maximize the infoNCE
based loss $\cL_{\text{ARL}}$:

\begin{equation}
\label{eq:Larl}
\cL_{\text{ARL}} = \mathbb{E}_{\mbx \sim \cX}
\left[\log \frac{e^{f(\mby_{l}, \mby^{\prime}_{l})}}{\frac{1}{N}
\sum_{m=1}^{N} e^{f(\mby_{l}, \mby^{\prime}_{m})}}
\right]
\end{equation}

For an input image $\mbx$
and two outputs $\mby$ and $\mby^{\prime}$. Let 
$\mby_l$ and $\mby^{\prime}_m$ correspond to $C$-dimensional
feature vectors at spatial positions $l$ and
$m$ in the first and second outputs respectively.
We start by creating $N$ pairs of feature vectors $(\mby_l, \mby_m^{\prime})$,
with one positive pair drawn from the joint distribution and $N-1$ negative pairs drawn from the marginals.
A positive pair is a pair of feature vectors corresponding to the same
spatial locations in the two outputs,
\ie, a pair $(\mby_l, \mby_m^{\prime})$ with $m=l$. The negatives 
are pairs $(\mby_l, \mby_m^{\prime})$ corresponding to two distinct
spatial positions $m\neq l$.
In practice, we also consider small displacements $\Omega$ (\cref{fig:overview}) when constructing positives.
Additionally, the negatives are generated from two distinct images, since
two feature vectors might share similar characteristics even with different spatial positions.
By maximizing \cref{eq:Larl}, we push the model $\cF$ to produce similar representations for the same
spatial location regardless of the applied ordering, so that the critic 
function $f$ is able to give high matching scores to the positive pairs and low matching to the negatives.
We follow \cite{infomax} and use separable
critics $f(\mby, \mby^{\prime})=\phi_1(\mby)^{\top} \phi_2(\mby^{\prime})$, where the functions 
$\phi_1 / \phi_2$ non-linearly transform the outputs to a higher vector space, and 
$f(\mby_l, \mby_m^{\prime})$ produces a scalar corresponding to a matching score between
the two representations at two spatial positions $l$ and $m$ of the two outputs.

Note that both losses $\cL_{\text{AC}}$ and $\cL_{\text{ARL}}$ can be applied
interchangeably for both objectives, a case we investigate in our experiments (\cref{sec:ablations}).
For $\cL_{\text{AC}}$, we can consider the clustering objective as an
intermediate task for learning useful representations.
For $\cL_{\text{ARL}}$, during inference, K-means \cite{JDH17} algorithm
can be applied over the outputs to obtain the cluster assignments.

\subsection{Model} \label{mdl}
The representation $\cF$ can be implemented in a general manner using three sub-parts, 
\ie, $\cF = h \circ g_{ar} \circ d$, with a feature extractor $h$, an autoregressive encoder
$g_{ar}$ and a decoder $d$. With such a formulation, the function $\cF$ is flexible and
can take different forms. With $h$ as an identity mapping, $\cF$ becomes a fully autoregressive
network, where we apply different orderings directly over
the inputs. Inversely, if $g_{ar}$
is an identity mapping, $\cF$ becomes a generic encoder-decoder network,
where $h$ plays the role of 
an encoder.
Additionally, $h$ can be a simple convolutional stem that plays
an important role in learning local features such as edges, or even multiple residual blocks \cite{resnet}
to extract higher representations. In this case, the orderings are applied over 
the hidden features using $g_{ar}$. $g_{ar}$ is similar to $h$,
containing a series of residual blocks,
with two main differences, the proposed masked convolutions are used,
and the batch normalization \cite{batchnorm}
layers are omitted to maintain the autoregressive dependency,
with an optional attention block.
The decoder $d$ can be a simple $\text{conv}1\times1$ to adapt the channels
to the number of cluster $K$, followed by bilinear upsampling and
a softmax operation for a clustering objective.
For representation learning, $d$ consists of two separable critics $\phi_1 / \phi_2$,
which are implemented as a series of $\text{conv}3\times3 - \text{BN} - \text{ReLU}$
and $\text{conv}1\times1$ for projecting to a higher dimensional space. See sup. mat. for
the architectural details. In the experimental section, we will investigate different versions of $\cF$.

\section{Experiments}

After stating the experimental setting, 
we start by presenting an extensive ablation study of the proposed method and its various parts.
We then compare the method to state-of-the-art approaches on unsupervised image 
segmentation.

\vspace{0.05in}
\noindent\textbf{Datasets.}
The experiments are conducted on the newly established and challenging baselines by \cite{IIC}.
Potsdam \cite{potsdam} with 8550 RGBIR satellite images of size $200 \times 200$,
of which 3150 are unlabeled. We experiment on both the 6-labels variant (roads and cars,
vegetation and trees, buildings and clutter) and Potsdam-3, a 3-label variant formed by
merging each of the pairs.
We also use COCO-Stuff \cite{cocostuff}, a dataset containing
\textit{stuff} classes. Similarly, we use a reduced version of
COCO-Stuff with 164k images and 15 coarse labels, reduced to 52k
by taking only images with at least 75\% stuff pixel.
In addition to COCO-Stuff-3 with only 3 labels, sky, ground and plants.

\vspace{0.05in}
\noindent\textbf{Evaluation Metrics.}
We report the pixel classification Accuracy (Acc).
For a clustering task, with a mismatch between the learned and ground truth clusters.
We follow the standard procedure and find the best one-to-one permutation
to match the output clusters to ground truth classes using the Hungarian algorithm \cite{hungarian}.
The Acc is then computed over the labeled examples.

\vspace{0.05in}
\noindent\textbf{Implementation details.}
The different variations of $\cF$ are trained using ADAM with a learning rate of $10^{-5}$
to optimize both objectives in \cref{eq:Lac,eq:Larl}. We train on  $200 \times 200$ crops
for Potsdam and  $128 \times 128$ for COCO.
The training is conducted on NVidia V100 GPUs, and implemented using the PyTorch framework \cite{pytorch}.
For more experimental details, see sup. mat.

\subsection{Ablation Studies} \label{sec:ablations}
\begin{figure}[t]
	\centering
	\includegraphics[width=0.8\linewidth]{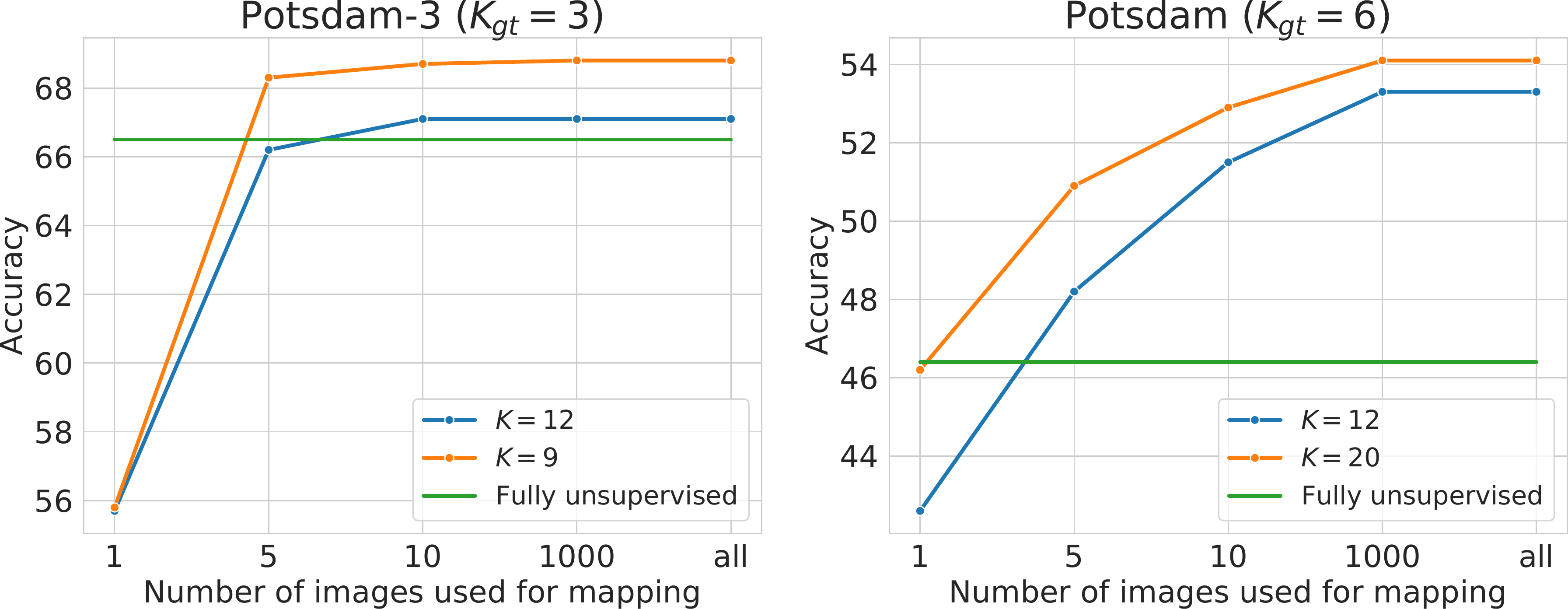}
	\caption{\textbf{Overclustering.} The Acc obtained when using a number of output clusters 
	greater than the number of ground truth classes $K>K_{gt}$. With variable number of images used
	to find the best many-to-one matching between the outputs and targets.}
	\label{fig:overclustering}
\end{figure}

We start by performing comprehensive ablation studies on the different
components and variations of the proposed method. 
\cref{tab:ablations,} and \cref{fig:overclustering} show the ablation results
for AC, and \cref{tab:arl} shows a comparison between AC and ARL,
analyzed as follows:

\begin{table*}[t]\centering\vspace{-3mm}
\hspace{2mm}
\subfloat[\textbf{Variation of $\cF$:} we compare different variants of 
the network $\cF$ using different feature extractors $f$ and
autoregressive encoders $g_{ar}$. The decoder $d$ is fixed.
\label{tab:ablation:fvariations}]{
\tablestyle{5pt}{1.}
\resizebox{0.5\columnwidth}{!}{%
\begin{tabular}{lll|x{22}x{22}x{33}}
\toprule
\multicolumn{3}{c|}{Network $\cF = h \circ g_{ar} \circ d$ } & & \\
& $h$ & $g_{ar}$ & POS & POS3 \\
\midrule
\multicolumn{3}{c|}{Random} & 28.5 & 38.2 \\
$\cF_1$ & Id 				& 5 Res. blocks		& 39.3 & 56.3 \\
$\cF_2$ & Stem 				& 5 Res. blocks   	& 46.4 & \textbf{66.4} \\
$\cF_3$ & Res. block 		& 4 Res. blocks   	& \textbf{47.9} & 64.5  \\
$\cF_4$ & 5 Res. blocks   	& Id   				& 35.1 & 63.4  \\
$\cF_5$ & ResNet-18 	    & Id   				& 40.7 & 51.9 \\
\toprule
\end{tabular}%
}
}\hspace{3mm}
\subfloat[\textbf{Number of orderings:}
we compare different sizes of the set $\cO$.
For $|\cO|=2$ and $|\cO|=4$ , we report the mean and std over 4 runs using
different possible pairs and quadruples respectively.
\label{tab:ablation:orderings}]{
\resizebox{0.37\columnwidth}{!}{%
\tablestyle{5pt}{1.}
\begin{tabular}{c|x{44}x{44}}
\toprule
$|\cO|$ & POS & POS3 \\
\midrule
2 & 43.2$\pm$2.19 & 59.5$\pm$5.12 \\
4 & 45.6$\pm$3.22 & 63.55$\pm$3.52 \\
8 & \textbf{46.4} & \textbf{66.4} \\
\toprule
\end{tabular}}
}

\subfloat[\textbf{Attention:}
we add a single attention block at a shallow level, 
and change the applied masks to get the desired orderings.
Output stride $= 4$ in this instance.
\label{tab:ablation:att}]{
\resizebox{0.5\columnwidth}{!}{%
\tablestyle{5pt}{1.}
\begin{tabular}{ccc|x{22}x{22}}
\toprule
\multicolumn{3}{c|}{Orderings} & & \\
Raster-Scan & Zigzag & Attention & POS & POS3 \\
\midrule
$\checkmark$ & $\times$ & $\times$ & 45.2 & 61.0 \\
$\checkmark$ & $\times$ & $\checkmark$ & 47.9 & 66.3 \\
$\times$ & $\checkmark$ & $\checkmark$ & 47.8 & \textbf{66.5} \\
$\checkmark$ & $\checkmark$ & $\checkmark$ & \textbf{49.3} & 65.4 \\
\toprule
\end{tabular}}
}
\hspace{10mm}
\subfloat[\textbf{Sampling of $o_i$:} we compare
different possible sampling procedures of the orderings $o_i$ during training.
\label{tab:ablation:sampling}]{
\resizebox{0.3\columnwidth}{!}{%
\tablestyle{5pt}{1.}
\begin{tabular}{c|x{22}x{22}}
\toprule
Sampling $o_i$ & POS & POS3 \\
\midrule
Random & 46.4 & \textbf{66.4} \\
No Rep. & 48.6 & 64.8 \\
Hard & \textbf{48.9} & 65.2 \\
\toprule
\end{tabular}}
}

\subfloat[\textbf{Transformations:}
we apply a given transformation to the inputs of the second
forward pass during a single training iteration.
\label{tab:ablation:trsf}]{
\resizebox{0.45\columnwidth}{!}{%
\tablestyle{5pt}{1.}
\begin{tabular}{cc|x{22}x{22}}
\toprule
Type & Transf. & POS & POS3 \\
\midrule
None & - & 46.4 & 66.4 \\
Photometric & Col. Jittering & 47.9 & 65.5 \\
Geometric & Flip & 46.7 & 68.0 \\
Geometric & Rot. & \textbf{48.5} & 68.3 \\
Geo. \& Pho. & All & \textbf{48.5} & \textbf{68.3} \\
\toprule
\end{tabular}}
}\hspace{10mm}
\subfloat[\textbf{Dropout:} we inspect the addition of 
dropout to the inner activations of a residual block.
\label{tab:ablation:dropout}]{
\resizebox{0.23\columnwidth}{!}{%
\tablestyle{5pt}{1.}
\begin{tabular}{c|x{22}x{22}}
\toprule
\textbf{p} & POS & POS3 \\
\midrule
0 & 46.4 & \textbf{66.4} \\
0.1 & \textbf{47.9} & 64.7 \\
0.2 & 46.9 & 65.1 \\
\toprule
\end{tabular}}
}
\vspace{.5em}
\caption{\textbf{AC Ablations.} Ablations studies conducted on Potsdam (POS) and Potsdam-3 (POS3) for Autoregressive Clusterings. We show 
the pixel classification accuracy (\%).}
\label{tab:ablations}
\vspace{-0.3in}
\end{table*}

\noindent\textbf{Variations of $\cF$.}
\cref{tab:ablation:fvariations} compares different variations of the network $\cF$.
With a fixed decoder $d$ (\ie, a $1\times1 \conv$ followed by bilinear
upsampling and softmax function), we adjust $h$ and $g_{ar}$ going from a
fully autoregressive model ($\cF_1$) to a normal decoder-encoder
network ($\cF_4$ and $\cF_5$). When using masked versions, we see an improvement over
the normal case, with up to 8 points for Potsdam, and to a lesser extent for Potsdam-3 where
the task is relatively easier with only three ground truth classes. 
When using a fully autoregressive model ($\cF_1$), and applying the orderings directly
over the inputs, maximizing the MI becomes much harder, and the model fails to
learn meaningful representations.
Inversely, when no masking is applied ($\cF_4$ and $\cF_5$), the task becomes
comparatively simpler, and we see a drop in performance. The best
results are obtained when applying the orderings over
low-level features ($\cF_2$ and $\cF_3$).
Interestingly, the unmasked versions yield results better than random, and perform
competitively with 3 output classes for Potsdam-3, validating the effectiveness of maximizing
the MI over small displacements $\mbu \in \Omega$. For the rest of the experiments we use $\cF_2$
as our model.

\noindent\textbf{Attention and different orderings.}
\cref{tab:ablation:att} shows the effectiveness of adding attention
blocks to our model. With a single attention block added at a shallow level,
we observe an improvement over the baseline, for both raster-scan and zigzag orderings,
and their combination, with up to 4 points for Potsdam.
In this case, given the quadratic complexity of attention, we used an output stride of 4.

\noindent\textbf{Data augmentations.}
For a given training iteration, we pass the same image two times through the network,
applying two different orderings at each forward pass.
We can, however, pass a transformed version of the image as the second input.
We investigate using photometric (\ie, color jittering) and geometric (\ie, rotations and H-flips) 
transformations. For geometric transformations, we bring the outputs back to
the input coordinate space before computing the loss.
Results are shown in \cref{tab:ablation:trsf}. As expected, we obtain relative improvements
with data augmentations, highlighting the flexibility of the approach.

\noindent\textbf{Dropout.}
To add some degree of stochasticity to the network, and as an additional regularization, 
we apply dropout to the intermediate activations within residual blocks of the network. 
\cref{tab:ablation:dropout} shows a small increase in Acc for Potsdam.

\noindent\textbf{Orderings.}
Until now, at each forward pass, we sample a pair of possible orderings with replacement
from the set $\cO$. With such a sampling procedure, we might end-up with
the same pair of orderings for a given training iteration. As an alternative, we investigate
two other sampling procedures. First, with no repetition (No Rep.), where we choose two distinct
orderings for each training iteration. Second, using hard sampling, choosing two orderings with opposite receptive 
fields (\eg, $o_1$ and $o_6$). \cref{tab:ablation:sampling} shows the obtained results.
We see 2 points improvement when using hard sampling for Potsdam. For simplicity, we use random sampling for the rest of the experiments.

Additionally, to investigate the effect of the number of orderings (\ie, the cardinality of $\cO$),
we compute the Acc over different choices and sizes of $\cO$. \cref{tab:ablation:orderings}
shows best results are obtained when using all 8 raster-scan orderings. Interestingly,
for some choices, we observe better results, which may be due to selecting orderings
that do not share any receptive fields, as the ones used in hard sampling.

\noindent\textbf{Overclustering.}
To compute the Acc for a clustering task using linear assignment,
the output clusters are chosen to match the ground truth classes $K = K_{gt}$. Nonetheless,
we can choose a higher number of clusters $K > K_{gt}$, and then find
the best many-to-one matching between the output clusters and ground truths based a given number of labeled examples. 
In this case, however, we are not in
a fully unsupervised case, given that we extract some information, although limited,
from the labels. \cref{fig:overclustering} shows that, even with a very limited
number of labeled examples used for mapping, we can obtain better results than
the fully unsupervised case.

\begin{table*}[t]\centering\vspace{-3mm}
{
\resizebox{0.275\columnwidth}{!}{%
\tablestyle{5pt}{1.}
\begin{tabular}{c|x{22}x{22}}
\toprule
\multicolumn{3}{c}{Clustering} \\
\midrule
Method & POS & POS3 \\
\midrule
Random CNN & 28.5 & 38.2 \\
AC & \textbf{46.4} & \textbf{66.4} \\
ARL & 45.1 & 57.1 \\
\bottomrule
\end{tabular}}\hspace{3mm}
}
{
\resizebox{0.275\columnwidth}{!}{%
\tablestyle{5pt}{1.}
\begin{tabular}{c|x{22}x{22}}
\toprule
\multicolumn{3}{c}{Linear Evaluation} \\
\midrule
Method & POS & POS3 \\
\midrule
AC & \textbf{23.7} & \textbf{41.4} \\
ARL & \textbf{23.7} & 38.5 \\
\bottomrule
\end{tabular}}\hspace{6mm}
}
{
\resizebox{0.275\columnwidth}{!}{%
\tablestyle{5pt}{1.}
\begin{tabular}{c|x{22}x{22}}
\toprule
\multicolumn{3}{c}{Non-Linear Evaluation} \\
\midrule
Method & POS & POS3 \\
\midrule
AC & \textbf{68.0} & \textbf{81.8} \\
ARL & 47.6 & 63.5 \\
\bottomrule
\end{tabular}}%
}
\vspace{.5em}
\caption{\textbf{Comparing ARL and AC.} We compare ARL and AC on a clustering task (left). And
investigate the quality of the learned representations by freezing the trained model, and reporting 
the test Acc obtained when training a linear (center) and non-linear (right) functions trained on the labeled training examples.}
\label{tab:arl}
\vspace{-0.2in}
\end{table*}

\noindent\textbf{AC and ARL}
To compare AC and ARL, we apply them interchangeably on both clustering and representation learning objectives.
In clustering, for ARL, after PCA Whitening, we apply K-means over the output features to get the cluster assignments.
In representation learning,
we evaluate the quality of the learned representations using both linear and non-linear separability
as a proxy for disentanglement, and as a measure of MI between representations and class
labels. \cref{tab:arl} shows the obtained results.

\noindent\textit{Clustering.} As expected, AC outperforms ARL on a clustering task, given that the clusters
are directly optimized by computing the exact MI during training.

\noindent\textit{Quality of the learned representations.}
Surprisingly, AC outperforms ARL on both linear and non-linear classifications. We hypothesize that 
unsupervised representation learning objectives that work well on image classification, fail in image segmentation
due to the dense nature of the task.
The model in this case needs to output distinct representations over pixels, rather than the whole
image, which is a harder task to optimize. This might also be due to using only a small number of features (\ie, $N$ pairs)
for each training iteration. 

\vspace{-0.1in}
\subsection{Comparison with the state-of-the-art}
\cref{tab:sota} shows the results of the comparison.
AC outperforms previous work,
and by a good margin for harder segmentation tasks with
a large number of output classes (\ie, Potsdam and COCO-Stuff), highlighting the effectiveness of maximizing the MI between
the different orderings as a training objective. We note that no regularization or data augmentation were used, 
and we expect that better results can be obtained by combining AC with other procedures as demonstrated in the ablation studies.
\begin{table}[t]
\centering
\setlength{\tabcolsep}{1pt}
\fontsize{8}{9}\selectfont 
\begin{tabular}{lc@{\hskip 0.1in}c@{\hskip 0.3in}c@{\hskip 0.1in}c}
\toprule
& COCO-Stuff-3  & COCO-Stuff & Potsdam-3 & Potsdam  \\
\midrule
Random CNN & 37.3  & 19.4  & 38.2 & 28.3 \\
K-means~\cite{scikit-learn} & 52.2 & 14.1 & 45.7 & 35.3\\
SIFT~\cite{lowe2004distinctive}& 38.1 & 20.2 & 38.2 & 28.5 \\
Doersch 2015~\cite{doersch} & 47.5 & 23.1 & 49.6  & 37.2 \\
Isola 2016~\cite{isola2015learning} & 54.0 & 24.3 & 63.9 & 44.9 \\
DeepCluster 2018~\cite{deepcluster} & 41.6 & 19.9 & 41.7 & 29.2 \\
IIC 2019~\cite{IIC} & 72.3 & 27.7 & 65.1 & 45.4 \\
\midrule
AC & \textbf{72.9} & \textbf{30.8} & \textbf{66.5} & \textbf{49.3} \\
\bottomrule
\end{tabular}
\vspace{0.05in}
\caption{\textbf{Unsupervised image segmentation.}
Comparison of AC with state-of-the-art methods on unsupervised segmentation.}
\label{tab:sota}
\vspace{-0.2in}
\end{table}

\section{Conclusion} 
We presented a novel method to create different \textit{views} of the inputs
using different %
\textit{orderings}, and showed
the effectiveness of maximizing the MI over these views for unsupervised image segmentation.
We showed that for image segmentation, optimizing over the discrete outputs by computing 
the exact MI works better for both clustering and unsupervised representation learning, due
to the dense nature of the task.
Given the simplicity and ease of adoption of the method, we hope that the proposed approach can be adapted 
for other visual tasks and used in future works.

\noindent \textbf{Acknowledgments.} Y. Ouali is supported by Randstad corporate research chair in collaboration with CentraleSupélec.
We would also like to thank Saclay-IA plateform of Université Paris-Saclay and Mésocentre computing center of CentraleSupélec and
École Normale Supérieure Paris-Saclay for providing the computational resources.

\bibliographystyle{splncs04}
\bibliography{paper}

\begin{thebibliography}{10}
\providecommand{\url}[1]{\texttt{#1}}
\providecommand{\urlprefix}{URL }
\providecommand{\doi}[1]{https://doi.org/#1}

\bibitem{amdim}
Bachman, P., Hjelm, R.D., Buchwalter, W.: Learning representations by
  maximizing mutual information across views. In: Advances in Neural
  Information Processing Systems. pp. 15509--15519 (2019)

\bibitem{becker1992self}
Becker, S., Hinton, G.E.: Self-organizing neural network that discovers
  surfaces in random-dot stereograms. Nature  \textbf{355}(6356),  161--163
  (1992)

\bibitem{attaugmentedconv}
Bello, I., Zoph, B., Vaswani, A., Shlens, J., Le, Q.V.: Attention augmented
  convolutional networks. In: Proceedings of the IEEE International Conference
  on Computer Vision. pp. 3286--3295 (2019)

\bibitem{bielski2019emergence}
Bielski, A., Favaro, P.: Emergence of object segmentation in perturbed
  generative models. In: Advances in Neural Information Processing Systems. pp.
  7256--7266 (2019)

\bibitem{cocostuff}
Caesar, H., Uijlings, J., Ferrari, V.: Coco-stuff: Thing and stuff classes in
  context. In: Proceedings of the IEEE Conference on Computer Vision and
  Pattern Recognition. pp. 1209--1218 (2018)

\bibitem{deepcluster}
Caron, M., Bojanowski, P., Joulin, A., Douze, M.: Deep clustering for
  unsupervised learning of visual features. In: Proceedings of the European
  Conference on Computer Vision (ECCV). pp. 132--149 (2018)

\bibitem{deepercluster}
Caron, M., Bojanowski, P., Mairal, J., Joulin, A.: Unsupervised pre-training of
  image features on non-curated data. In: Proceedings of the IEEE International
  Conference on Computer Vision. pp. 2959--2968 (2019)

\bibitem{redo}
Chen, M., Arti{\`e}res, T., Denoyer, L.: Unsupervised object segmentation by
  redrawing. In: Advances in Neural Information Processing Systems. pp.
  12705--12716 (2019)

\bibitem{psnail}
Chen, X., Mishra, N., Rohaninejad, M., Abbeel, P.: Pixelsnail: An improved
  autoregressive generative model. In: International Conference on Machine
  Learning. pp. 864--872 (2018)

\bibitem{sparsetransformers}
Child, R., Gray, S., Radford, A., Sutskever, I.: Generating long sequences with
  sparse transformers. arXiv preprint arXiv:1904.10509  (2019)

\bibitem{doersch}
Doersch, C., Gupta, A., Efros, A.A.: Unsupervised visual representation
  learning by context prediction. In: Proceedings of the IEEE International
  Conference on Computer Vision. pp. 1422--1430 (2015)

\bibitem{deepkmeans}
Fard, M.M., Thonet, T., Gaussier, E.: Deep $ k $-means: Jointly clustering with
  $ k $-means and learning representations. arXiv preprint arXiv:1806.10069
  (2018)

\bibitem{federici2020learning}
Federici, M., Dutta, A., Forr{\'e}, P., Kushman, N., Akata, Z.: Learning robust
  representations via multi-view information bottleneck. arXiv preprint
  arXiv:2002.07017  (2020)

\bibitem{potsdam}
Gerke, M.: Use of the stair vision library within the isprs 2d semantic
  labeling benchmark (vaihingen)  (2014)

\bibitem{MADE}
Germain, M., Gregor, K., Murray, I., Larochelle, H.: Made: Masked autoencoder
  for distribution estimation. In: International Conference on Machine
  Learning. pp. 881--889 (2015)

\bibitem{imrot}
Gidaris, S., Singh, P., Komodakis, N.: Unsupervised representation learning by
  predicting image rotations. arXiv preprint arXiv:1803.07728  (2018)

\bibitem{pascalvoc}
Girshick, R., Donahue, J., Darrell, T., Malik, J.: Rich feature hierarchies for
  accurate object detection and semantic segmentation. In: Proceedings of the
  IEEE conference on computer vision and pattern recognition. pp. 580--587
  (2014)

\bibitem{xavier}
Glorot, X., Bengio, Y.: Understanding the difficulty of training deep
  feedforward neural networks. In: Proceedings of the thirteenth international
  conference on artificial intelligence and statistics. pp. 249--256 (2010)

\bibitem{associativeclustering}
Haeusser, P., Plapp, J., Golkov, V., Aljalbout, E., Cremers, D.: Associative
  deep clustering: Training a classification network with no labels. In: German
  Conference on Pattern Recognition. pp. 18--32. Springer (2018)

\bibitem{clustering}
Hartigan, J.A.: Direct clustering of a data matrix. Journal of the american
  statistical association  \textbf{67}(337),  123--129 (1972)

\bibitem{resnet}
He, K., Zhang, X., Ren, S., Sun, J.: Deep residual learning for image
  recognition. In: Proceedings of the IEEE conference on computer vision and
  pattern recognition. pp. 770--778 (2016)

\bibitem{kANMI}
He, Z., Xu, X., Deng, S.: k-anmi: A mutual information based clustering
  algorithm for categorical data. Information Fusion  \textbf{9}(2),  223--233
  (2008)

\bibitem{infomax}
Hjelm, R.D., Fedorov, A., Lavoie-Marchildon, S., Grewal, K., Bachman, P.,
  Trischler, A., Bengio, Y.: Learning deep representations by mutual
  information estimation and maximization. arXiv preprint arXiv:1808.06670
  (2018)

\bibitem{IMSAT}
Hu, W., Miyato, T., Tokui, S., Matsumoto, E., Sugiyama, M.: Learning discrete
  representations via information maximizing self-augmented training. In:
  Proceedings of the 34th International Conference on Machine Learning-Volume
  70. pp. 1558--1567. JMLR. org (2017)

\bibitem{segsort}
Hwang, J.J., Yu, S.X., Shi, J., Collins, M.D., Yang, T.J., Zhang, X., Chen,
  L.C.: Segsort: Segmentation by discriminative sorting of segments. In:
  Proceedings of the IEEE International Conference on Computer Vision. pp.
  7334--7344 (2019)

\bibitem{batchnorm}
Ioffe, S., Szegedy, C.: Batch normalization: Accelerating deep network training
  by reducing internal covariate shift. arXiv preprint arXiv:1502.03167  (2015)

\bibitem{isola2015learning}
Isola, P., Zoran, D., Krishnan, D., Adelson, E.H.: Learning visual groups from
  co-occurrences in space and time. arXiv preprint arXiv:1511.06811  (2015)

\bibitem{IIC}
Ji, X., Henriques, J.F., Vedaldi, A.: Invariant information clustering for
  unsupervised image classification and segmentation. In: Proceedings of the
  IEEE International Conference on Computer Vision. pp. 9865--9874 (2019)

\bibitem{JDH17}
Johnson, J., Douze, M., J{\'e}gou, H.: Billion-scale similarity search with
  gpus. arXiv preprint arXiv:1702.08734  (2017)

\bibitem{wnet}
Kanezaki, A.: Unsupervised image segmentation by backpropagation. In: 2018 IEEE
  international conference on acoustics, speech and signal processing (ICASSP).
  pp. 1543--1547. IEEE (2018)

\bibitem{kanezaki2018unsupervised}
Kanezaki, A.: Unsupervised image segmentation by backpropagation. In: 2018 IEEE
  international conference on acoustics, speech and signal processing (ICASSP).
  pp. 1543--1547. IEEE (2018)

\bibitem{adam}
Kingma, D.P., Ba, J.: Adam: A method for stochastic optimization. arXiv
  preprint arXiv:1412.6980  (2014)

\bibitem{hungarian}
Kuhn, H.W.: The hungarian method for the assignment problem. Naval research
  logistics quarterly  \textbf{2}(1-2),  83--97 (1955)

\bibitem{NADE}
Larochelle, H., Murray, I.: The neural autoregressive distribution estimator.
  In: Proceedings of the Fourteenth International Conference on Artificial
  Intelligence and Statistics. pp. 29--37 (2011)

\bibitem{lowe2004distinctive}
Lowe, D.G.: Distinctive image features from scale-invariant keypoints.
  International journal of computer vision  \textbf{60}(2),  91--110 (2004)

\bibitem{NWJ}
Nguyen, X., Wainwright, M.J., Jordan, M.I.: Estimating divergence functionals
  and the likelihood ratio by convex risk minimization. IEEE Transactions on
  Information Theory  \textbf{56}(11),  5847--5861 (2010)

\bibitem{pcnn}
Van~den Oord, A., Kalchbrenner, N., Espeholt, L., Vinyals, O., Graves, A.,
  et~al.: Conditional image generation with pixelcnn decoders. In: Advances in
  neural information processing systems. pp. 4790--4798 (2016)

\bibitem{prnn}
Oord, A.v.d., Kalchbrenner, N., Kavukcuoglu, K.: Pixel recurrent neural
  networks. arXiv preprint arXiv:1601.06759  (2016)

\bibitem{cpc}
Oord, A.v.d., Li, Y., Vinyals, O.: Representation learning with contrastive
  predictive coding. arXiv preprint arXiv:1807.03748  (2018)

\bibitem{imtransformer}
Parmar, N., Vaswani, A., Uszkoreit, J., Kaiser, L., Shazeer, N., Ku, A., Tran,
  D.: Image transformer. In: Dy, J., Krause, A. (eds.) Proceedings of the 35th
  International Conference on Machine Learning. Proceedings of Machine Learning
  Research, vol.~80, pp. 4055--4064. PMLR, Stockholmsmässan, Stockholm Sweden
  (10--15 Jul 2018)

\bibitem{pytorch}
Paszke, A., Gross, S., Chintala, S., Chanan, G., Yang, E., DeVito, Z., Lin, Z.,
  Desmaison, A., Antiga, L., Lerer, A.: Automatic differentiation in pytorch
  (2017)

\bibitem{scikit-learn}
Pedregosa, F., Varoquaux, G., Gramfort, A., Michel, V., Thirion, B., Grisel,
  O., Blondel, M., Prettenhofer, P., Weiss, R., Dubourg, V., et~al.:
  Scikit-learn: Machine learning in python. Journal of machine learning
  research  \textbf{12}(Oct),  2825--2830 (2011)

\bibitem{pcnnpp}
Salimans, T., Karpathy, A., Chen, X., Kingma, D.P.: Pixelcnn++: Improving the
  pixelcnn with discretized logistic mixture likelihood and other
  modifications. arXiv preprint arXiv:1701.05517  (2017)

\bibitem{CMC}
Tian, Y., Krishnan, D., Isola, P.: Contrastive multiview coding. arXiv preprint
  arXiv:1906.05849  (2019)

\bibitem{onMImaximization}
Tschannen, M., Djolonga, J., Rubenstein, P.K., Gelly, S., Lucic, M.: On mutual
  information maximization for representation learning. arXiv preprint
  arXiv:1907.13625  (2019)

\bibitem{attalluneed}
Vaswani, A., Shazeer, N., Parmar, N., Uszkoreit, J., Jones, L., Gomez, A.N.,
  Kaiser, {\L}., Polosukhin, I.: Attention is all you need. In: Advances in
  neural information processing systems. pp. 5998--6008 (2017)

\bibitem{nonlocal}
Wang, X., Girshick, R., Gupta, A., He, K.: Non-local neural networks. In:
  Proceedings of the IEEE conference on computer vision and pattern
  recognition. pp. 7794--7803 (2018)

\bibitem{atgan}
Zhang, H., Goodfellow, I., Metaxas, D., Odena, A.: Self-attention generative
  adversarial networks. arXiv preprint arXiv:1805.08318  (2018)

\end{thebibliography}


\begin{thebibliography}{1}
\providecommand{\url}[1]{\texttt{#1}}
\providecommand{\urlprefix}{URL }
\providecommand{\doi}[1]{https://doi.org/#1}

\bibitem{cocostuff}
Caesar, H., Uijlings, J., Ferrari, V.: Coco-stuff: Thing and stuff classes in
  context. In: Proceedings of the IEEE Conference on Computer Vision and
  Pattern Recognition. pp. 1209--1218 (2018)

\bibitem{xavier}
Glorot, X., Bengio, Y.: Understanding the difficulty of training deep
  feedforward neural networks. In: Proceedings of the thirteenth international
  conference on artificial intelligence and statistics. pp. 249--256 (2010)

\bibitem{IIC}
Ji, X., Henriques, J.F., Vedaldi, A.: Invariant information clustering for
  unsupervised image classification and segmentation. In: Proceedings of the
  IEEE International Conference on Computer Vision. pp. 9865--9874 (2019)

\bibitem{adam}
Kingma, D.P., Ba, J.: Adam: A method for stochastic optimization. arXiv
  preprint arXiv:1412.6980  (2014)

\bibitem{cpc}
Oord, A.v.d., Li, Y., Vinyals, O.: Representation learning with contrastive
  predictive coding. arXiv preprint arXiv:1807.03748  (2018)

\end{thebibliography}

\clearpage
\section*{Supplementary Material}
\addcontentsline{toc}{section}{Appendices}
\renewcommand{\thesection}{\Alph{section}}
\setcounter{section}{0}

\noindent In this supplementary material, we provide architectural details, hyperparameters settings,
further discussions about the loss functions
and the masked convolutions. We also provide some qualitative results and implementation details.

\section{Architectural details}
\cref{tab:stem,tab:resblock,tab:decoder,tab:sepcritics} 
present the building blocks of the representation function $\cF$. Specifically, 
we describe the architecture of the convolutional stem, the residual blocks, the decoder for AC and the separable critics used for ARL.

\begin{table}[h]
\centering
\fontsize{9}{11}\selectfont 
    \begin{tabular}{l@{\hskip 0.6in}c}
    \hline
    \multicolumn{2}{c}{Convolutional Stem}\\
    		Layer  & Output size \\
    \Xhline{2\arrayrulewidth}
        \textit{Input} & $3\times H \times W$ \\
        Conv $3 \times 3$ & $64\times H \times W$ \\
        Batch Norm - ReLU & $64\times H\times W$ \\
        Max Pool $3 \times 3$, $s = 2$ & $64 \times H/2 \times W/2$  \\ 
    \hline
    \end{tabular}
    \vspace{0.1in}
    \caption{Convolutional Stem for an output stride of $2$. For an output stride of $4$,
    we use a Conv $3 \times 3$ with stride of $2$, yielding an output of size $64 \times H/4 \times W/4$.
    For the fully autoregressive case, the Batch Norm and Max pool are omitted, and the Max pool is replaced
    with a strided masked convolution.}
    \label{tab:stem}
    \vspace{-0.5in}
\end{table}

\begin{table}[ht!]
\centering
\fontsize{9}{11}\selectfont 
    \begin{tabular}{l@{\hskip 0.6in}c}
    \hline
    \multicolumn{2}{c}{Decoder}\\
            Layer  & Output size \\
    \Xhline{2\arrayrulewidth}
        \textit{Input} & $C\times H/2 \times W/2$ \\
        Conv $1 \times 1$ & $K\times H \times W$ \\
        Bilinear Interpolation & $K\times H\times W$ \\
        Softmax & $K\times H\times W$\\ 
    \hline
    \end{tabular}
    \vspace{0.1in}
    \caption{Decoder used for a clustering objective. In this case, we have
    an output stride of 2 and $K$ clusters.}
    \label{tab:decoder}
    \vspace{-0.4in}
\end{table}

\begin{table}[ht!]
\centering
\fontsize{9}{11}\selectfont 
    \begin{tabular}{l@{\hskip 0.6in}c}
    \hline
    \multicolumn{2}{c}{Separable Critics}\\
            Layer  & Output size \\
    \Xhline{2\arrayrulewidth}
        \textit{Input} & $C\times H \times W$ \\
        Conv $1 \times 1$ - ReLU & $2C\times H \times W$ \\
        Conv $1 \times 1$ - ReLU & $2C\times H \times W$ \\
        Conv $1 \times 1$ applied to the input & $2C\times H \times W$ \\
        Residual Connection + Batch Norm & $2C\times H \times W$ \\
    \hline
    \end{tabular}
    \vspace{0.1in}
    \caption{Separable critics used for representation learning to non-linearly project
    the outputs to a higher vector space.}
    \label{tab:sepcritics}
    \vspace{-0.1in}
\end{table}

\begin{table}[ht!]
\centering
\fontsize{9}{11}\selectfont 
    \begin{tabular}{l@{\hskip 0.6in}c}
    \hline
    \multicolumn{2}{c}{Residual Block}\\
    		Layer  & Output size \\
    \Xhline{2\arrayrulewidth}
        \textit{Input} & $C\times H \times W$ \\
        Conv $3 \times 3$ - ReLU & $2C\times H \times W$ \\
        Conv $1 \times 1$ - ReLU & $2C\times H\times W$ \\
        Zero padding of the input to $2C$ & $2C\times H \times W$  \\ 
        Residual Connection & $2C\times H \times W$ \\
    \hline
        Conv $1 \times 1$ - ReLU & $2C\times H \times W$ \\
        Conv $1 \times 1$ - ReLU & $2C\times H\times W$ \\
        Residual Connection & $2C\times H \times W$ \\
    \hline
        Conv $1 \times 1$ - ReLU & $2C\times H \times W$ \\
        Conv $1 \times 1$ - ReLU & $2C\times H\times W$ \\
        Residual Connection & $2C\times H \times W$ \\
    \hline
    \end{tabular}
    \vspace{0.1in}
    \caption{Architecture of residual blocks, for residual blocks used in the
    autoregressive encoder $g_{ar}$, normal
    convolutions are replaced with masked ones.}
    \label{tab:resblock}
    \vspace{-0.2in}
\end{table}

\section{Hyperparameters}

We discuss the hyperparameters used in our experiments. We note that we have noticed that
the network is very sensitive to
the initialization. In our case, we initialize the parameters using Xavier initialization \cite{xavier},
and noticed a somehow more stable results with such an instantiation scheme. The optimizer of choice
is Adam \cite{adam} with the default parameters ($\beta_1 = 0.9$ and $\beta_2 = 0.999$). The rest of the 
hyperparamters used are detailed in \cref{tab:hyper}.

\noindent For the transformations applied in Paper Section 4.1, we used color jittering where we randomly change the brightness, hue,
contrast and saturation of an image up to $10\%$ for photometric transformations.
For geometric transformation, we apply random horizontal flips and random rotations
by multiples of 90 degrees.

\begin{table}
\centering
\vspace{-0.15in}
\fontsize{9}{13}\selectfont 
    \begin{tabular}{lc@{\hskip 0.1in}c@{\hskip 0.1in}c@{\hskip 0.1in}c}
    \hline
    Parameter & COCO-stuff 3 & COCO-stuff & Potsdam-3 & Potsdam \\
    \Xhline{2\arrayrulewidth}
    LR  & $4.10^{-5}$ & $6.10^{-6}$ & $10^{-6}$ & $4.10^{-5}$ \\
    Batch size & 60 & 60 & 30 & 30 \\
    Crop size & 128$\times$128 & 128$\times$128 & 200$\times$200 & 200$\times$200 \\
    Rescale factor & 0.33 & 0.33 & No rescal. & No rescal. \\
    Output stride & 4 & 4 & 2 & 2 \\
    Num. of displacements for $\cL_{AC}$ & 10 & 10 & 10 & 10 \\
    Attention & False & False & True & True \\
    \hline
    \end{tabular}
    \vspace{0.1in}
    \caption{Hyperparameters used for training per dataset.}
    \label{tab:hyper}
    \vspace{-0.2in}
\end{table}

\section{Loss functions}
In this section, we will go into more details about the loss functions introduced in Paper Section 3.2.
For a given unlabeled input $\mbx \sim \cX$,
and two outputs $\mby \sim \cF(\mbx; o_i)$ and $\mby^{\prime} \sim \cF(\mbx; o_j)$ with
two valid orderings $(o_i, o_j) \in \cO$,
the training objective is to maximize the MI between the two encoded variables:
\begin{equation}
\label{eq:mainobj}
\max _{\cF} I(\mby; \mby^{\prime})
\end{equation}

\subsection{Autoregressive Clustering $\cL_{AC}$}
To see the benefits of maximizing \cref{eq:mainobj} for a clustering objective, we expand the objective as
the difference between two entropy terms:
\begin{equation}
\label{eq:midef}
I(\mby; \mby^{\prime})=H(\mby)-H(\mby|\mby^{\prime})
\end{equation}
By such a formulation, we can see that maximizing the MI involves maximizing the entropy and minimizing the conditional entropy.
The compromise between these two terms help us avoid both degenerate and trivial solutions.
For degenerate solution, where the model $\cF$ outputs uniform distributions over all of the pixels, not assigning
any cluster to any pixel, the entropy $H(\mby)$ in this case is maximized,
however the second term $H(\mby|\mby^{\prime})$ is also 
maximized, since the outputs are not deterministic and there is
no predictability of the second output from the first.
Inversely, with trivial solutions, where all of the pixel are assigned to the same cluster. 
the second output $\mby^{\prime}$ is totally deterministic from the first, and the conditional 
entropy $H(\mby|\mby^{\prime})$ is minimized, yet, the entropy $H(\mby)$ is also minimized and we fail to maximize the MI.
By balancing the maximization of the first term and the minimization of the second, we are more likely to end-up
with the correct assignments, than if we only maximized the entropy.

Given that the two outputs are generated using the same input and two different orderings, there is 
a strong statistical dependency between them. In this case,
$\mby \sim \cF(\mbx; o_i)$ and $\mby^{\prime} \sim \cF(\mbx; o_j)$ are
dependent and we compute the joint probability $p(\mby,\mby^{\prime})$
as a matrix of size $K \times K$:
\begin{equation}
\label{eq:joint}
p(\mby,\mby^{\prime})= \cF(\mbx; o_i)^T \cF(\mbx; o_j)
\end{equation}

In practice we also marginalize over the batch, with an input $\mbx$ of shape $B \times 3 \times H \times W$
as a batch of $B$ input images.
Let $\mbx_i$ correspond to the i-th image in the batch $\mbx$ of $B$ images. In this
case the joint probability is computed as follows:
\begin{equation}
\label{eq:joint}
p(\mby,\mby^{\prime})= \frac{1}{B} \sum_{i=1}^B \cF(\mbx_i; o_i)^T \cF(\mbx_i; o_j)
\end{equation}

Additionally, following \cite{IIC}, we also compute the joint probability over small possible displacements $\mbu\in \Omega$.
Let the input $\mbx^{(\mbu)}$ correspond to shifting the input $\mbx$ by $\mbu$ pixels (\ie, zero padding and cropping).
In such a case, we also need to marginalize over all possible displacements $\mbu$ as follows:
\begin{equation}
\label{eq:joint}
p(\mby,\mby^{\prime})= \frac{1}{B} \frac{1}{|\Omega|} \sum_{i=1}^B \sum_{\mbu \in \Omega} \cF(\mbx_i; o_i)^T \cF(\mbx_i^{(\mbu)}; o_j)
\end{equation}

Finally, by summing over the rows and columns of $p(\mby,\mby^{\prime})$, we can compute the marginals,
and then the MI:
\begin{equation}
\label{eq:mikl}
I(\mby, \mby^{\prime}) = D_{\mathrm{KL}}(p(\mby, \mby^{\prime})
\| p(\mby) p(\mby^{\prime}))
\end{equation}

\begin{figure}
\centering
\includegraphics[width=\linewidth]{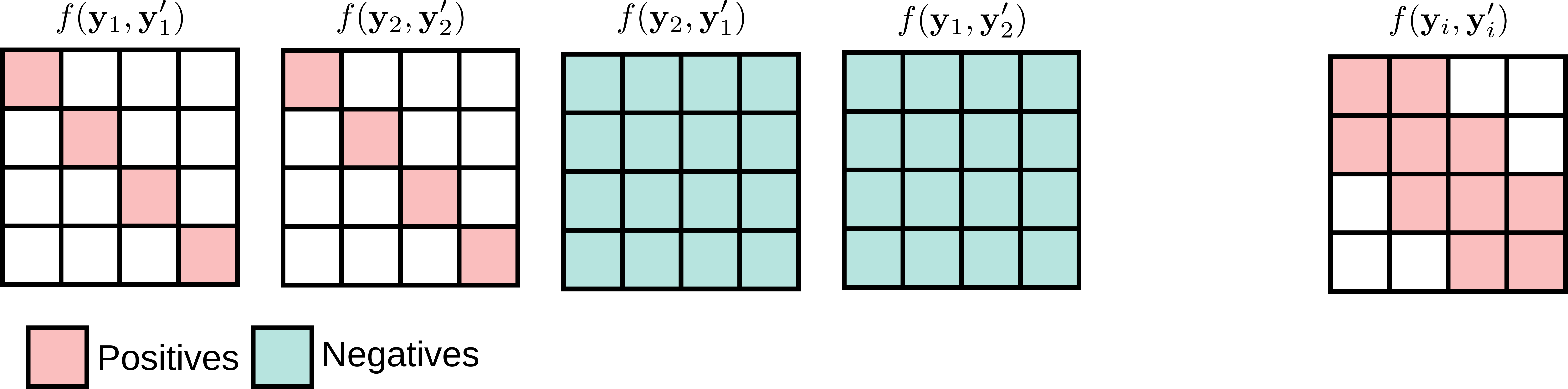}
\vspace{-0.1in}
\caption{\textit{Left}: Examples of positive and negative pairs for $B = 2$ and $HW = 4$. $\mby_i$ refers to the 
i-th element of the output $\mby$ corresponding to the i-th image in the input batch. \textit{Right}: Examples of positive pairs
with possible displacements $\Omega = \{-1, 0, 1\}.$}
\label{fig:larl}
\vspace{-0.2in}
\end{figure}

\subsection{Autoregressive Representation Learning $\cL_{AC}$}
For unsupervised representation learning objective, we maximize the infoNCE \cite{cpc} as a lower
bound of MI over the continuous outputs:
\begin{equation}
\label{eq:Larl}
\cL_{\text{ARL}} = 
\log \frac{e^{f(\mby_{l}, \mby^{\prime}_{l})}}{\frac{1}{N}
\sum_{m=1}^{N} e^{f(\mby_{l}, \mby^{\prime}_{m})}}
\end{equation}

The goal of \cref{eq:Larl} is to push the network $\cF$ to produce similar
features between the two outputs $\mby$ and $\mby^\prime$ at the same spatial locations, so that
the critic is able to give high scores between two feature vectors ($\mby_{l}, \mby^{\prime}_{m}$)
at the same spatial position $m = l$, and low scores for feature vectors from distinct spatial position $m \neq l$
or from two distinct images.
To compute the loss in \cref{eq:Larl}, we need to create a set of positive and negative pairs.
With a batch of images $\mbx$ of shape $B \times 3 \times H \times W$, we 
generate two outputs $\mby$ and $\mby^\prime$ of shape $B \times C \times H \times W$, with $C$-dimensional
output feature maps. In this case
the output of the critic $f(\mby, \mby^\prime) = \phi_1(\mby)^{\top} \phi_2(\mby^{\prime})$
is a matrix of shape $BHW \times BHW$. To construct the positive and 
negative pairs, we reshape the scoring matrix as $B^2$ matrices of shape $HW$, in this case
the positives are the diagonals of each matrix from the same images with a given shift $\mbu \in \Omega$.
The negatives are all of the possible combination across the matrices from distinct images. See \cref{fig:larl} for
an illustration for $B = 2$ and $HW = 4$. Note that
we avoid using the same image to construct negative pairs, and only construct
them across images, given that even with distinct spatial positions,
it is very likely that two feature vectors share similar characteristics.

\section{Receptive fields}

\begin{figure}
\centering
\includegraphics[width=0.8\linewidth]{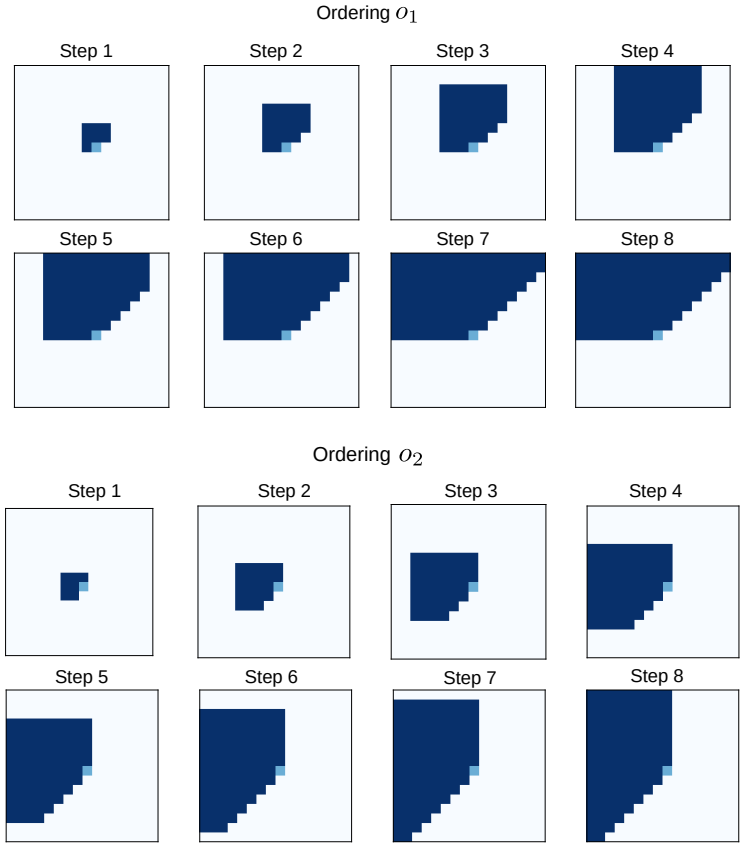}
\vspace{-0.05in}
\caption{Examples of the growing receptive field of pixel \crule[cyan]{0.2cm}{0.2cm} for two orderings; $o_1$ and $o_2$,
over 8 consecutive applications of masked convolutions to get the correct orderings. As expected, after enough convolutions,
and with the correct shift, we can construct the desired ordering. Note that in both cases we have a significant number
of pixels in the blind spots, which can be accessed using an attention block. In this case,
we use $\conva$ with $\shifta$ and $\shiftb$.}
\label{fig:recep}
\vspace{-0.1in}
\end{figure}

To further illustrate how a given ordering $o_i$ is constructed, we present a toy example where we 
plot the receptive field of a given pixel at the center of an image of size $16 \times 16$. After each
application of a masked convolution with the corresponding shift, we compute the gradient of the target pixel
and plot the non-zero values in blue, which correspond to the receptive field of the
target pixel. The results are illustrated in \cref{fig:recep}.

\begin{figure}[h]
  \centering
\includegraphics[width=0.7\linewidth]{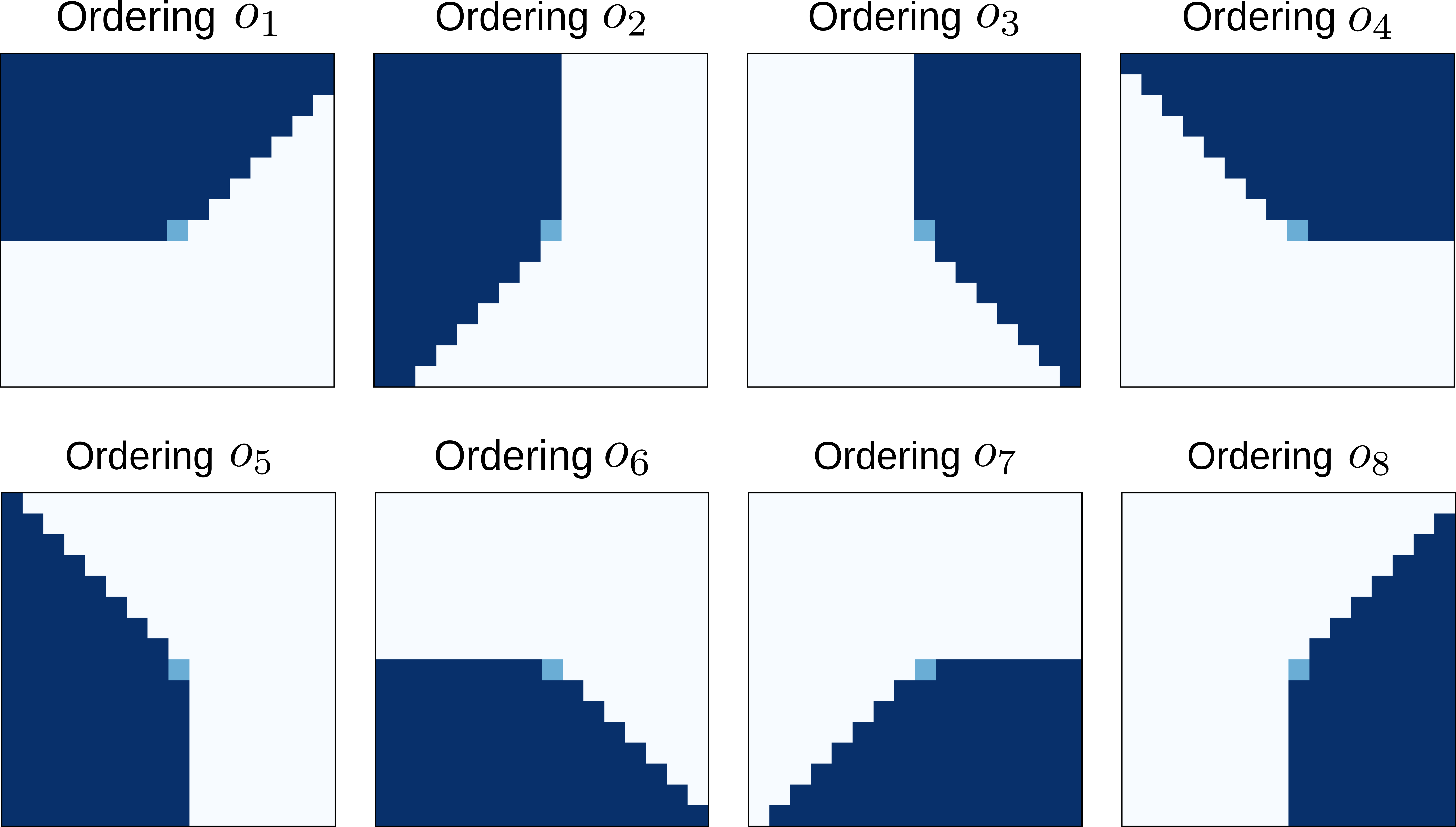}
\caption{The resulting receptive fields with the various raster-scan type orderings}
\label{fig:recep_orderings}
\vspace{-0.3in}
\end{figure}

\noindent\textbf{Orderings.} For a given pair of distinct orderings, the resulting dependencies and receptive fields of the two outputs will be different even if the applied orderings are quite similar. It is however likely that the two outputs share some overlap in their receptive fields, but such an overlap is small and helps reduce the difficulty of the task. An illustration of the resulting receptive fields for a given pixel using raster-scan orderings is shown in \cref{fig:recep_orderings}.

\section{Qualitative Results}
\begin{figure}[t]
\centering
\includegraphics[width=0.9\linewidth]{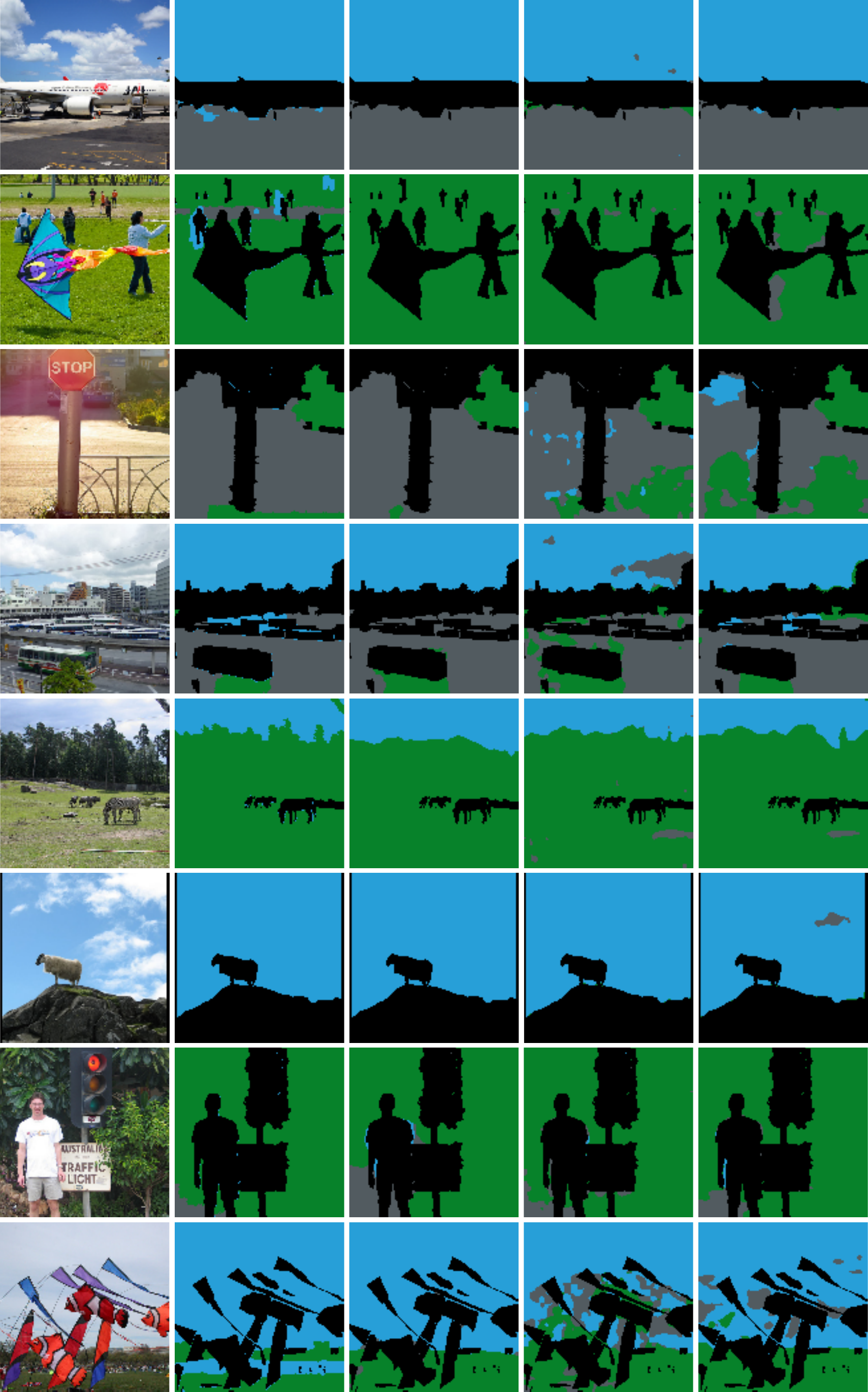}

\vspace{-0.15in}
\caption{Qualitative Results from COCO-Stuff 3
\cite{cocostuff,IIC} \textit{test} set.}
\label{fig:qualitative}
\vspace{-0.1in}
\end{figure}

\cref{fig:qualitative} shows qualitative results of Autoregressive Clustering (AC) on COCO-stuff 3 \textit{test} set,
in addition to linear and non-linear evaluations, where the model trained for AC is frozen and then the corresponding layers
are added on top of the decoder, that are then trained on the \textit{train} set.
Surprisingly, even if the accuracy with linear and non-linear evaluations is higher, we see that qualitatively, the fully unsupervised
method gives slightly better results. This might be due to the dense nature of image segmentation,
where the prediction at a given pixel is very dependent of its neighbors, and we lose this locality with linear evaluation,
given that we consider each pixel as a standalone data point. This is similar to what we observed with
ARL where we optimize the representations at each spatial location separately. Note that we have noticed
some minor annotation errors in the ground truths that might be due to the conversion done by \cite{IIC},
these are very minor and can be overlooked.

\begin{figure}[t]
\centering
\includegraphics[width=0.9\linewidth]{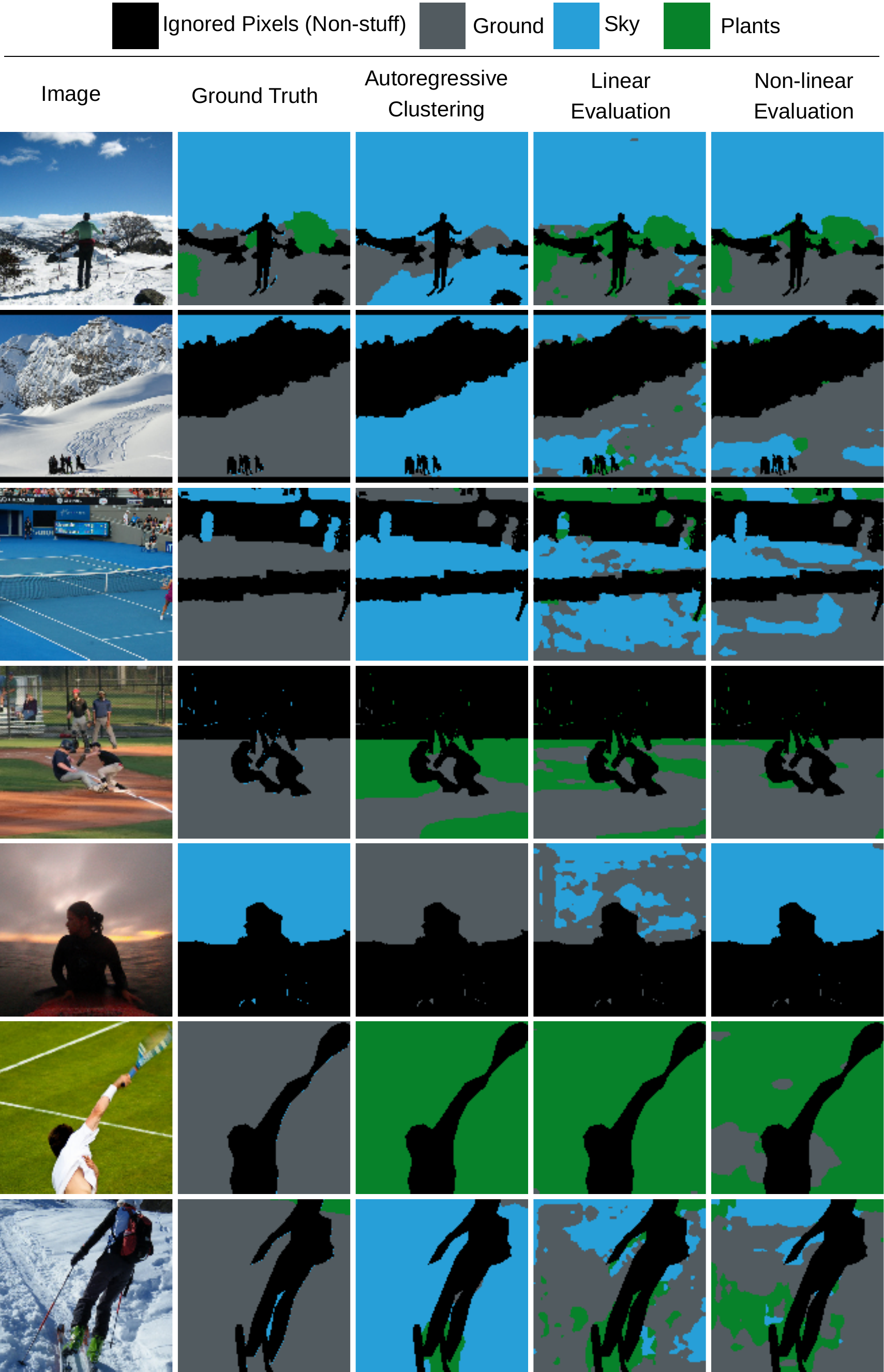}

\caption{Failure Cases for Autoregressive Clustering from COCO-Stuff 3
\cite{cocostuff,IIC} \textit{test} set.}
\label{fig:failures}
\vspace{-0.1in}
\end{figure}

\vspace{0.15in}
\noindent We also present some examples where AC fails in \cref{fig:failures}. We observe that the model is very dependent on the appearance
and colors for making the predictions. However, in special cases, like tennis courts with grass or asphalt floors, the model predicts
green or sky classes, and the correct prediction is ground.
This can be overcome with additional data augmentations like color jittering, or in case where
a limited amount of labeled examples are available, the model can be fine-tunned to correct such mistakes.
We already see some slight improvements with linear and non-linear evaluations.

\begin{figure}[t]
\centering
\includegraphics[width=0.9\linewidth]{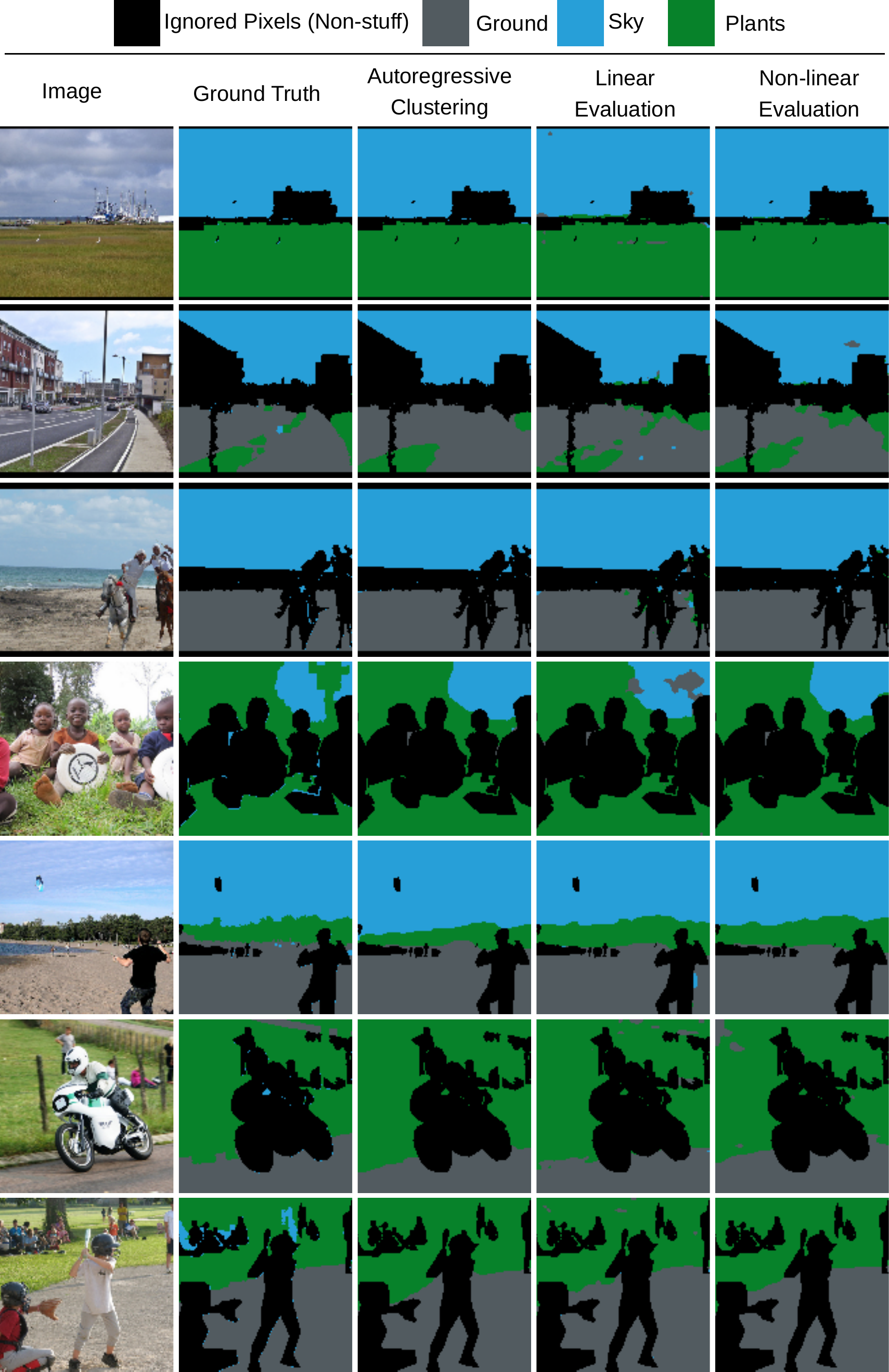}
\vspace{-0.15in}
\vspace{-0.1in}
\end{figure}

\end{document}